\pdfoutput=1
\documentclass[11pt]{article}

\usepackage[final]{ACL2023}

\usepackage{times}
\usepackage{latexsym}
\usepackage{booktabs, multirow} % for borders and merged ranges
\usepackage{soul}% for underlines
\usepackage{xcolor,colortbl} % for cell colors
\usepackage{changepage,threeparttable} % for wide tables
\usepackage{amsmath}
\usepackage{graphicx}
\usepackage{svg}
\usepackage{hyperref}

\usepackage{subcaption}
\usepackage[utf8]{inputenc}

\usepackage{microtype}

\usepackage{inconsolata}

\title{Data-Efficient Hate Speech Detection via Cross-Lingual Nearest Neighbor Retrieval with Limited Labeled Data}
\author{Faeze Ghorbanpour$^{1,2}$\qquad Daryna Dementieva$^{1}$\qquad Alexander Fraser$^{1, 2}$ \vspace{.2cm}\\ 
$^{1}$School of Computation, Information and Technology, TU Munich \\
$^{2}$Munich Center for Machine Learning (MCML)\\
\vspace{.1cm} {\tt \small faeze.ghorbanpour@tum.de, daryna.dementieva@tum.de} %alexander.fraser@tum.de}
}

\begin{document}
\maketitle
\begin{abstract}
Considering the importance of detecting hateful language, labeled hate speech data is expensive and time-consuming to collect, particularly for low-resource languages. 
Prior work has demonstrated the effectiveness of cross-lingual transfer learning and data augmentation in improving performance on tasks with limited labeled data.
To develop an efficient and scalable cross-lingual transfer learning approach, we leverage nearest-neighbor retrieval to augment minimal labeled data in the target language, thereby enhancing detection performance. 
Specifically, we assume access to a small set of labeled training instances in the target language and use these to retrieve the most relevant labeled examples from a large multilingual hate speech detection pool. 
We evaluate our approach on eight languages and demonstrate that it consistently outperforms models trained solely on the target language data. 
Furthermore, in most cases, our method surpasses the current state-of-the-art. 
%Notably, our approach is highly data-efficient, requiring only 200 to 2,000 retrieved instances while maintaining superior performance. 
Notably, our approach is highly data-efficient, retrieving as small as 200 instances in some cases while maintaining superior performance.
Moreover, it is scalable, as the retrieval pool can be easily expanded, and the method can be readily adapted to new languages and tasks. 
We also apply maximum marginal relevance to mitigate redundancy and filter out highly similar retrieved instances, resulting in improvements in some languages.\footnote{The pool and code will be made publicly available.}$^{,}$\footnote{\textcolor{red}{This paper contains examples of hateful language.}}

\end{abstract}

\section{Introduction}
% Why labeled data is important and costly for hate speech (data-efficient and limited target language data)
Hate speech, \textit{abusive language targeting specific groups} \citep{rottger2020hatecheck}, is a global issue. However, most detection advancements focus on English due to the abundance of labeled datasets \citep{poletto2021resources, yin2021towards}. In contrast, languages like Spanish, French, and Italian, though not low-resource for other tasks, lack annotated hate speech datasets \citep{poletto2021resources}, limiting model effectiveness in detecting and addressing hate speech.

Collecting and annotating data for low-resource languages is an effective solution, especially for capturing linguistic and cultural nuances in hate speech \citep{pelicon2021investigating, aluru2021deep}. As \citet{rottger2022data} state, having some labeled data in the target language is crucial for model effectiveness. However, while obtaining more data can improve performance, this requires paying high annotation costs \citep{elsherief-etal-2021-latent} and exposes annotators to harmful content \citep{alemadi2024emotional}.

% \begin{figure}[t] % Force placement at the top
%   \centering
%   \includegraphics[width=\linewidth]{figures/Picture1.png} % Replace with your image file
%   \caption{Overview of the proposed method. Given a small number of examples from a target language, we search a large pool of multilingual data to find closely related instances. We then combine the retrieved instances with the target language data and train a multilingual model on them for hate speech detection.
%   }
%   \label{fig:main_structure}
% \end{figure}

Transfer learning, especially from high-resource languages like English, helps mitigate data scarcity and improve detection performance \citep{bigoulaeva2022addressing, 10.1016/j.eswa.2023.121115}. However, the choice of source tasks and languages remains crucial. Some languages are useful for specific target languages due to cultural similarities \citep{zhou2023cross}, and certain source tasks may be more useful for particular target tasks \citep{rottger2022data, antypas2023robust}.

Training on all available hate speech datasets may seem beneficial, but is often inefficient, computationally costly, and does not guarantee better performance \citep{caselli-etal-2020-feel}. It can introduce redundancy, dataset-specific biases, and annotation inconsistencies, leading to overfitting \citep{wiegand2019detection, 10.1145/3232676}. Moreover, this approach lacks scalability, requiring frequent retraining for new datasets \citep{vidgen-etal-2021-introducing}.

\begin{figure*}[t] % Force placement at the top
  \centering
  \includegraphics[width=\linewidth]{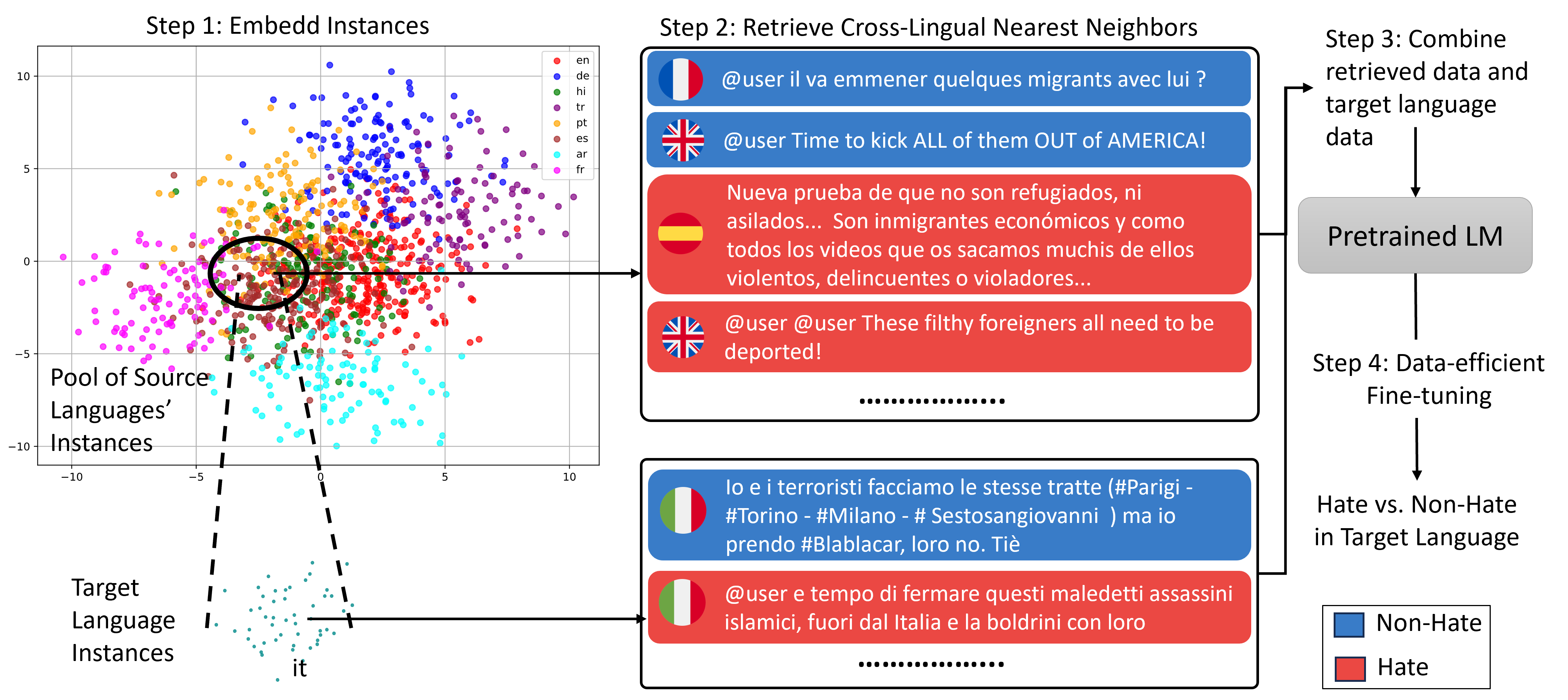} 
  \caption{Overview of the proposed method. Given a small number of examples from a target language, we search in a large pool of multilingual data for closely related instances. We then combine the retrieved instances with the target language data and train a multilingual model on them for hate speech detection.
  }
  \label{fig:main_structure}
\end{figure*}

To address the mentioned problems, we propose a novel method based on cross-lingual nearest-neighbor retrieval. Our approach, pictured in Figure \ref {fig:main_structure}, retrieves a minimal yet relevant set of instances and integrates them with the target language training set for fine-tuning. Specifically, we embed all available instances from fourteen tasks using a multilingual sentence embedding model to create a pool of hate speech detection samples. A retrieval system selects the most relevant instances from the multilingual pool based on their distance to the target language training set. These retrieved instances are then combined with the target training data to fine-tune a language model (LM).

This solution addresses multiple challenges. First, retrieving from a multilingual pool eliminates the need to manually select source tasks or languages. Second, it enhances efficiency by retrieving only a small number of relevant samples, avoiding redundancy through distance-based retrieval. Third, it ensures scalability, as the multilingual pool can be easily expanded with new datasets and languages. Our method improves cross-lingual transfer learning by considering linguistic and semantic similarities in hate speech across languages.

We evaluate the proposed method on eight languages, including German, French, Spanish, Italian, Portuguese, Hindi, Arabic, and Turkish, simulating a scenario where only a limited number of training examples (ranging from 10 to 2,000) are available. Fine-tuning on a combination of retrieved data and the target language training set significantly outperformed fine-tuning solely on the target training set across all languages. Further, our method outperformed the state-of-the-art work in most languages while fine-tuning with fewer samples. To refine the retrieved data, we also experiment with applying maximum marginal relevance (MMR)~\citep{carbonell1998use} to remove highly similar instances, leading to improved performance in some languages. Our contributions are as follows: % TODO To Be Completed! 

\begin{itemize}
\item We propose a novel, efficient, and scalable method for enhancing limited labeled hate speech datasets by retrieving cross-lingual samples using a retrieval system.
\item We evaluate our method on eight languages, demonstrating consistently higher performance compared to training solely on the target language training set.
\item Our approach is particularly effective in extremely low-resource settings, with fewer than 50 labeled instances, achieving up to 10\% F1-macro improvement in some cases. 
\end{itemize}

% 

% 

%

% Our contribution:
% covering 8 non-english languages with limited target data simulations
% is data efficient -> retrieving and fine-tuning on much less data
% embedding pool is scalable, and new datasets and other languages can be added
\section{Related Work}
\paragraph{Hate Speech Detection with Limited Labeled Data: }
Hate speech violates human rights, disrupts social peace, incites violence, and promotes discrimination in all societies, regardless of language. Detecting it is crucial to prevent conflict and protect mental health and societal safety \citep{wilson2019digital, narula2024comprehensive}. Most datasets and research efforts focus on English \cite{toraman-etal-2022-large, kennedy2020constructing, vidgen-etal-2021-learning}, while languages like Spanish, Portuguese, or Ukrainian have very limited resources. According to the \href{https://hatespeechdata.com/}{Hate Speech Dataset Catalogue}, these languages each have only one available dataset with fewer than 5,000 samples, which are also restricted in context and domain \cite{basile-etal-2019-semeval, fortuna-etal-2019-hierarchically, dementieva-etal-2024-toxicity}. Due to limited resources, recent research increasingly leverages other languages to improve hate speech detection in low-resource settings.

\noindent
\textbf{Cross-lingual Transfer Learning for Hate Speech Detection: }
Cross-lingual transfer learning has been widely studied in NLP, showing that models trained on high-resource languages can improve performance in low-resource languages \citep{parovic2023cross, muraoka2023cross, pham2024unibridge}. This makes it a promising approach for hate speech detection in low-resource settings. Early methods used multilingual embeddings for zero-shot and few-shot transfer from resource-rich to resource-poor languages \cite{aluru2020deep, pamungkas-patti-2019-cross}. Also, \citet{bigoulaeva2021cross} and \citet{arango2021crosslingualhatespeechdetection} utilized bilingual embeddings to transfer knowledge from high-resource languages, showing promising results even without labeled data in target languages but mainly benefiting closely related languages. 
%\citet{nozza-2021-exposing} highlighted the challenges of zero-shot transfer, emphasizing the need for language-specific considerations.
%https://arxiv.org/pdf/2303.16828
%Some studies aim to improve cross-lingual performance. 

Data augmentation strategies, including cross-lingual paraphrasing or translation-based methods, have also been shown to alleviate data scarcity \citep{pamungkas2021joint, beddiar2021data}, but these approaches are often constrained by the availability and quality of translation resources. \citet{roychowdhury2023data} employed data augmentation with EasyMixup and reframed the task as textual entailment, achieving improvements but still relying on potentially noisy augmented data.
% Hate Speech Detection in Limited Data Contexts Using Synthetic Data Generation (https://dl.acm.org/doi/10.1145/3625679) TODO
\citet{hashmi2025metalinguist}, \citet{gharoun2024meta}, and \citet{mozafari2022cross} use meta-learning approaches specialized for bilingual contexts. While effective, these methods require extensive labeled bilingual data, are complex to implement and train, and often demand substantial computational resources, making them less scalable.
%\citet{singh-thakur-2024-generalizable} proposed MultiFED, a federated learning framework, which improves performance without centralizing data and preserving privacy, but requires complex coordination across distributed sources.

\citet{rottger2022data} showed that minimal target-language data and initial English fine-tuning improve performance. However, selecting an appropriate intermediate English task is challenging and language-dependent. Building on this, \citet{goldzycher2023evaluating} uses an intermediate natural language inference (NLI) task, which adds training steps and requires more computation.
Unlike prior approaches, our method eliminates the need for large-scale target-language annotation, intermediate tasks, or translation resources. Directly leveraging semantic similarity at the instance level enables effective transfer with minimal target data and avoids costly cross-lingual training pipelines.

\noindent
\textbf{Retrieval-based and Instance attribution Fine-tuning methods:}
Prior work has shown that cross-task retrieval-based data can improve generalization in LMs \citep{guu2020retrieval, khandelwal2019generalization}. \citet{shi2022knn} applied retrieval to classification tasks via heuristic label mapping, whereas we fine-tune directly on nearest neighbors. \citet{das-khetan-2024-deft} introduces data-efficient fine-tuning through unsupervised core-set selection, showing strong results in monolingual text-editing tasks. However, this method is not designed for cross-lingual transfer and depends on clustering quality. 

Our approach is similar to \citet{lin2022unsupervised} and \citet{ivison2022data} in using nearest neighbor retrieval and further fine-tuning, but is uniquely applied to multilingual datasets and leverages labeled hate speech data. % and then combines retrieved and labeled target instances, enabling effective transfer.
Our method uses instance attribution, identifying relevant training examples for a data point, unlike prior work \citep{pruthi2020estimating, han2022orca}, which used gradient-based instance attribution to interpret neural network predictions. Our neighbor identification approach is simpler as it avoids gradient computations and reliance on labels, and is applied in a multilingual, low-resource setting.

\section{Methodology}
Building on a large pool of labeled multilingual hate speech data, our core hypothesis is that certain instances in this pool are more relevant to a given target language than others. For each target language, we assume access to a small amount of labeled data. The goal is to identify a relevant subset of source data that, when used for training, yields better performance.
Initially, we employ an embedding mode (\emph{Embedder}) to encode instances from multiple source languages. We then use a retrieval module (\emph{Indexer}) to index the resulting embedding vectors and construct a pool of multilingual hate speech detection instances.

When detecting hate speech in a low-resource target language, the objective is to fine-tune an LM for effective and efficient classification, as depicted in Figure~\ref{fig:main_structure}. We begin by embedding the target language instances using the same embedding model. The retrieval module (\emph{Retriever}) then searches the pool to find the nearest neighbors of the target instances. We combine the retrieved data with the target data and use the combined set to fine-tune (\emph{Fine-tuner}) an LM to classify them as Hate or Non-Hate. Each module is described below.

% To detect hate speech in a new target language with limited labeled data, we aim to fine-tune a language model for effective and efficient classification in that language. We begin by embedding the new language instances using an embedding module. Next, a retrieval module searches a data pool to find the nearest neighbors of these instances. Finally, we combine the retrieved neighbors with the new language data and use this combined set for fine-tuning.

% Finally, we apply Maximum Marginal Relevance (MMR) to eliminate the most similar embeddings, ensuring that we use the most relevant yet diverse data for training. 
% In the following subsections, we will go through the modules individually. 

%For simplicity, we illustrate the approach using a single source language. In practice, however, our methodology incorporates multiple source languages—eight in total. 
\subsection{Embedder and Indexer}
Assume a source language\footnote{For clarity, we describe the approach using a single source language. In practice, however, our methodology incorporates multiple source languages—eight in total.} \( A \) with a set of \( n \) text instances \( X^s = \{x^s_1, x^s_2, \ldots, x^s_n\} \) and corresponding labels \( Y^s \in \{0, 1\} \), where 1 indicates hate speech and 0 indicates non-hate speech. The objective of this module is to project the input texts into a vector space \( V^s = \{v^s_1, v^s_2, \ldots, v^s_n\} \), where each vector \( v^s_i \) is obtained by applying an embedding function: \( v^s_i = \text{embedding}(x^s_i) \).
These embeddings are then passed to the retrieval module, which indexes the vectors to enable efficient similarity search. This indexed embedding space serves as the foundation for retrieving relevant instances. % during downstream fine-tuning tasks.

\subsection{Retriever}
Consider a target language \( B \) with a limited set of labeled data \( X^t = \{x^t_1, x^t_2, \ldots, x^t_m\} \), where \( m \ll n \) 
(m and n denote the number of target and source language instances, respectively.),
and a label set \( Y^t \in \{0, 1\} \), where 1 denotes hate speech and 0 denotes non-hate speech (the same label set as the source language). Similar to the source language, we apply the embedding module to convert the target language instances into a numerical vector space \( V^t = \{v^t_1, v^t_2, \ldots, v^t_m\} \), where each vector is computed as \( v^t_i = \text{embedding}(x^t_i) \).

The retrieval module is then employed to find relevant samples from the pool using a nearest neighbor search. Specifically, we want to retrieve a total of \( R \) instances from the source pool based on \textit{Euclidean distance} between the embedded target vectors \( V^t \) and the source vectors \( V^s \). The distance between an embedded target instance \( v^t_i \) and a source instance \( v^s_j \) is calculated as:

\[
\text{dist}(v^t_i, v^s_j) = \|v^t_i - v^s_j\|_2 = \sqrt{\sum_{k=1}^{d} (v^t_{i,k} - v^s_{j,k})^2}
\]

Where \( d \) is the dimensionality of the embedding space.
We then select the top \( k \) nearest neighbors for each \( v^t_i \), and define the full retrieval set as:

\[
\mathcal{R} = \bigcup_{i=1}^{m} \text{TopK}(v^t_i, V^s, k)
\]

where \( \text{TopK}(v'_i, V^s, k) \) denotes the set of \( k \) source vectors in \( V^s \) with the lowest distance to \( v^t_i \). The set \( \mathcal{R} \) contains up to \( m \times k \) total retrieved instances. We then map the vectors in \( \mathcal{R} \) back to their corresponding original texts using the retrieval index (\( X^s_r=\{x^s_{r_1}, x^s_{r_2}, x^s_{r_3}, \ldots, x^s_{r_R}\} \)). Finally, we apply deduplication to remove exact textual duplicates.
%and ensure a diverse yet relevant support set. 
If the final count of unique instances falls short of \( R \), the retrieval process continues until the desired number is reached.

\subsection{Fine-tuner}
In the fine-tuning module, we combine the retrieved texts (\( X^s_r \)) with the training data from the target language (\( X^t \)). The combined dataset is then used to fine-tune a pre-trained LM (\( \mathcal{M} \)) to perform binary classification.
Since the source and target tasks share the same label space, i.e., \( Y^s, Y^t = \{0, 1\} \), where 0 denotes non-hate and 1 denotes hate, joint training of the fine-tuned model on the combined source and target data is well-defined and coherent.
We define the final training set as \( \mathcal{D} = X^s_r \cup X^t \). The model is fine-tuned by minimizing the binary cross-entropy loss:

\begin{multline*}
\mathcal{L} = - \frac{1}{|\mathcal{D}|} \sum_{(x, y) \in \mathcal{D}} \Big[ 
y \log \mathcal{M}(x) \\
+ (1 - y) \log \left(1 - \mathcal{M}(x)\right) \Big]
\end{multline*}

\section{Experimental Setup}

\subsection{Datasets}
% We use six large-scale English hate speech detection datasets as source tasks and evaluate the method on eight non-English hate speech detection tasks. 
% We select datasets that (a) contain explicit labels for hate, and (b) use a definition of hate speech for annotation that aligns with \citet{rottger2022data}.

We use six large-scale English hate speech detection datasets as well as eight non-English ones. 
These datasets were selected based on two criteria: (a) the presence of explicit hate speech labels, and (b) the use of annotation guidelines that align with, or are closely related to, the definition of hate speech adopted in this study.
The English datasets are: \textit{Dyn21\_en} \citep{vidgen-etal-2021-learning}, \textit{Fou18\_en} \citep{founta2018large}, \textit{Ken20\_en} \citep{kennedy2020contextualizing}, \textit{HateXplain} \citep{mathew2021hatexplain}, \textit{Implicit\_hate} \citep{elsherief-etal-2021-latent}, and \textit{Xdomain\_en} \citep{toraman-etal-2022-large}.

The non-English target tasks are: \textit{Bas19\_es} \citep{basile-etal-2019-semeval}, \textit{For19\_pt} \citep{fortuna-etal-2019-hierarchically}, \textit{Has21\_hi} \citep{mandl2021overview}, \textit{Our19\_ar} and \textit{Our19\_fr} \citep{ousidhoum-etal-2019-multilingual}, \textit{San20\_it} \citep{sanguinetti2020haspeede}, \textit{Xdomain\_tr} \citep{toraman-etal-2022-large}, and \textit{Gahd24\_de} \citep{goldzycher-etal-2024-improving}. The two-character language codes in dataset names indicate the language of the task. More information about datasets is provided in Appendix~\ref{sec:datasets_details}.

Although all datasets are embedded and included in the shared retrieval pool, we ensure that, for each non-English target task, instances from the same language are excluded from retrieval. This guarantees that the target language data remains unseen during its own retrieval process. Additionally, we exclude \textit{Dyn21\_en} when the target task is \textit{Gahd24\_de} because the latter includes translations from the former. We also exclude \textit{Xdomain\_en} when the target task is \textit{Xdomain\_tr}, as both originate from the same source.  
After constructing the multilingual pool, we obtain approximately 265{,}671 instances, of which 37.15\% are labeled as hateful. The majority of data in the pool is English (66.99\%), Turkish (17.0\%), and German (3.84\%).
%While this imbalance exists, it does not diminish the effectiveness of leveraging cross-lingual similarities, particularly among typologically close languages, as demonstrated in the evaluation section. 

% We further evaluated the model on out-of-domain evaluation sets of hate check and hate day? TODO

\begin{table*}[!htp]\centering
\scriptsize
\begin{tabular}{p{0.5cm}p{0.35cm}p{0.35cm}p{0.35cm}p{0.35cm}p{0.35cm}p{0.45cm}p{0.00001cm}p{0.35cm}p{0.35cm}p{0.35cm}p{0.35cm}p{0.35cm}p{0.45cm}p{0.00001cm}p{0.35cm}p{0.35cm}p{0.35cm}p{0.35cm}p{0.35cm}p{0.45cm}}\toprule
TRAIN&\multicolumn{6}{c}{San20\_it} & &\multicolumn{6}{c}{Ous19\_ar} & &\multicolumn{6}{c}{Ous19\_fr} \\\cmidrule{2-7}\cmidrule{9-14}\cmidrule{16-21}
SIZE &Mono &20 &200 &2{,}000 &20{,}000 &Röttger & &Mono &20 &200 &2{,}000 &20{,}000 &Röttger & &Mono &20 &200 &2{,}000 &20{,}000 &Röttger \\\midrule
20 &54.25 &63.20 &\underline{66.76} &\textbf{67.06} &60.06 &64.96 & &51.67 &\underline{57.63} &\textbf{63.23} &61.73 &59.47 &60.52 & &47.26 &47.21 &\underline{52.68} &53.93 &\textbf{55.05} &52.93 \\
50 &65.71 &67.20 &68.42 &\textbf{71.65} &69.44 &69.10 & &52.13 &59.36 &\underline{\textbf{66.65}} &66.31 &64.51 &65.76 & &47.87 &48.29 &\underline{52.19} &52.97 &\textbf{55.60} &54.15 \\
200 &72.81 &72.46 &\textbf{72.83} &72.41 &72.50 &71.56 & &67.97 &\underline{67.98} &\textbf{69.18} &67.35 &65.47 &66.61 & &51.93 &51.54 &\underline{54.06} &\textbf{55.80} &53.63 &53.76 \\
500 &74.18 &75.39 &\textbf{75.29} &74.53 &66.09 &73.69 & &66.54 &\underline{68.95} &\textbf{69.47} &69.28 &65.54 &67.60 & &51.91 &\underline{53.30} &52.84 &\textbf{55.51} &55.31 &53.39 \\
2{,}000 &76.40 &69.27 &\textbf{78.36} &77.57 &76.95 &77.07 & &66.91 &69.52 &69.77 &\textbf{70.15} &68.27 &67.07 & &51.84 &53.51 &53.13 &53.30 &\textbf{54.74} &52.89 \\\cmidrule{2-7}\cmidrule{9-14}\cmidrule{16-21}
AVG &66.53 &67.68 &71.00 &\textbf{71.06} &68.55 &70.47 & &59.82 &63.94 &\textbf{66.82} &66.41 &65.20 &65.41 & &49.72 &50.56 &53.12 &54.05 &\textbf{54.84} &53.88 \\\midrule
&\multicolumn{6}{c}{Bas19\_es} & &\multicolumn{6}{c}{For19\_pt} & &\multicolumn{6}{c}{Xdomain\_tr} \\\cmidrule{2-7}\cmidrule{9-14}\cmidrule{16-21}
20 &49.91 &54.37 &59.72 &\underline{62.52} &63.08 &\textbf{66.52} & &48.09 &49.72 &\underline{64.92} &\textbf{68.57} &68.03 &67.68 & &55.43 &66.58 &67.08 &70.14 &\underline{\textbf{75.87}} &69.80 \\
50 &61.85 &60.93 &64.37 &65.59 &64.30 &\textbf{70.36} & &60.25 &59.26 &\underline{67.01} &67.06 &\textbf{69.35} &66.51 & &72.24 &75.92 &77.50 &\textbf{78.85} &70.60 &75.12 \\
200 &72.36 &72.22 &71.77 &71.23 &70.67 &\textbf{75.27} & &66.91 &69.69 &\underline{70.33} &70.20 &\textbf{71.07} &68.10 & &81.63 &81.61 &82.61 &\textbf{83.04} &82.61 &82.19 \\
500 &77.14 &78.01 &77.09 &77.79 &67.67 &\textbf{78.76} & &69.95 &69.72 &70.84 &70.04 &\textbf{71.05} &69.22 & &85.05 &84.93 &85.09 &84.92 &83.88 &\textbf{85.34} \\
2{,}000 &81.08 &80.62 &80.50 &80.65 &81.02 &\textbf{82.04} & &\textbf{72.70} &72.39 &72.66 &71.72 &72.22 &71.61 & &88.53 &87.39 &88.00 &87.39 &77.48 &\textbf{88.84} \\\cmidrule{2-7}\cmidrule{9-14}\cmidrule{16-21}
AVG &65.52 &67.27 &69.53 &70.53 &68.69 &\textbf{72.97} & &61.66 &62.85 &68.18 &69.55 &\textbf{69.69} &68.39 & &73.58 &76.78 &78.88 &79.66 &77.58 &\textbf{80.27} \\\midrule
&\multicolumn{6}{c}{Gahd24\_de} & &\multicolumn{6}{c}{Has21\_hi} & & & & & & & \\\cmidrule{2-7}\cmidrule{9-14}
20 &44.99 &50.52 &\underline{58.15} &59.08 &57.48 &\textbf{59.82} & &46.87 &\underline{47.34} &51.03 &\underline{53.68} &\textbf{55.37} &54.92 & & & & & & & \\
50 &57.85 &54.53 &60.30 &60.47 &61.02 &\textbf{62.57} & &46.87 &48.39 &\underline{53.36} &52.26 &\textbf{55.78} &54.77 & & & & & & & \\
200 &65.80 &\underline{\textbf{66.95}} &66.15 &66.24 &65.97 &64.25 & &52.20 &55.83 &54.65 &\underline{56.80} &56.02 &\textbf{57.47} & & & & & & & \\
500 &66.56 &69.78 &69.68 &\textbf{70.45} &61.80 &67.02 & &56.20 &56.94 &\underline{57.66} &57.88 &\textbf{59.55} &57.96 & & & & & & & \\
2{,}000 &73.77 &\textbf{79.19} &78.79 &77.82 &77.90 &72.42 & &57.14 &58.19 &60.22 &\textbf{60.50} &59.65 &58.01 & & & & & & & \\\cmidrule{2-7}\cmidrule{9-14}
AVG &60.06 &62.98 &64.77 &\textbf{65.36} &64.18 &64.23 & &50.96 &52.44 &55.10 &56.25 &\textbf{57.05} &56.70 & & & & & & & \\
\bottomrule
\end{tabular}
\caption{ F1-macro scores across eight languages, comparing our method with the \texttt{Mono} and \texttt{Röttger}. Results are reported for target training sizes of 20, 50, 200, 500, and 2{,}000. Columns represent the number of retrieved instances. \texttt{AVG} denotes the mean over 12 training sizes. The best result for each language and training size is in \textbf{bold}. Retrieved results that outperform the next larger \texttt{Mono} training size are \underline{underlined}.
}\label{tab:evaluation_1}
\end{table*}

\subsection{Models}
For embedding the text instances, we utilized the \texttt{BAAI/bge-m3} multilingual encoder model \citep{chen-etal-2024-m3} using the Sentence Transformers library \citep{reimers-2020-multilingual-sentence-bert}. This model generates 1024-dimensional vector representations for each input text. We use the FAISS library \citep{douze2024faiss, johnson2019billion} to index dense vectors and perform a similarity search. For retrieval, we adopt the Hierarchical Navigable Small World (HNSW) algorithm \citep{malkov2018efficient} as an efficient approximation of the k-nearest neighbor search.
Throughout all our experiments for the classification model, we fine-tune and evaluate XLM-T \citep{barbieri2021xlm} using the HuggingFace Transformers library \citep{wolf2020transformers}. XLM-T is a variant of XLM-R \citep{conneau2019unsupervised}, further pre-trained on 198 million multilingual Twitter posts to better capture social media language patterns.  
%We compared XLM\-R, XLM\-T, and mDeBERTa \citep{he2021debertav3} on our limited labeled data and found that XLM-T performed better and covered more languages.
Further details on hyperparameters and experimental settings are provided in Appendix~\ref{sec:models_details}.

\subsection{Evaluation Details}
%As mentioned earlier, we index and store all embeddings of instances from all datasets in the shared retrieval pool. However, when performing retrieval and fine-tuning for a specific target language, we exclude any instances from that language to ensure it remains unseen during training.
%Following the setup of \citet{rottger2022data}, 
We simulate low-resource conditions by using 12 different training subset sizes per each non-English language: 10, 20, 30, 40, 50, 100, 200, 300, 400, 500, 1,000, and 2,000 examples. For each subset size, we run experiments with $5$ random seeds. 
%In each seed, a different set of training instances was sampled from the target language’s training split. 
Across all experiments, we use a fixed validation set of 500 examples and a test set of 2,000 examples for each target language\footnote{For Arabic and French, smaller dataset sizes limited the test sets to 1,000 and 1,500 samples, respectively.}. 
We only use the \textit{training split} of the target language for retrieval and fine-tuning. The test set remains entirely unseen throughout the process to ensure evaluation integrity and is kept fixed across all experiments.

The performance comparison is based on the F1-macro metric. We compare our method to the common practice of fine-tuning solely on the target training set, referred to as \texttt{Mono}. We also compare against the approach by \citet{rottger2022data}, which performs intermediate fine-tuning on three English hate speech datasets (20{,}000 instances each) to identify the most effective source task and then fine-tunes on the target language training set. We report the best result among the three as \texttt{Röttger}.

%We initially evaluated our method on the HateCheck functional test set \citep{rottger-etal-2022-multilingual, rottger2020hatecheck} but observed a strong correlation between this evaluation task and \textit{Dyn21\_en} across all languages. Specifically, when the number of retrieved instances from \textit{Dyn21\_en} was higher than from other tasks, the model performed better. This trend aligns with findings in the baseline paper by \citet{rottger2022data}, where intermediate fine-tuning on \textit{Dyn21\_en} yielded consistently superior results. These observations raised concerns about potential data contamination in this evaluation setting, so we ultimately chose not to include the results (further information is in Appendix~D).

\section{Results}
Table~\ref{tab:evaluation_1} presents a subset of the main results, showing five representative training sizes (20, 50, 200, 500, 2{,}000) out of the 12 evaluated across eight languages. The \texttt{AVG} column reflects the average performance over all 12 training sizes. The \texttt{Mono} and \texttt{Röttger} columns show the performance of those methods on the target language’s test set. Columns labeled 20, 200, 2{,}000, and 20{,}000 indicate the number of retrieved instances used from the multilingual pool. 
Further results are in Appendix~\ref{sec:full_results}.

In all languages, retrieving \textbf{as few as 20 instances} for fine-tuning already outperforms the \texttt{Mono} setting, indicating the effectiveness of our proposed method and the value of cross-lingual data. This is particularly promising for target tasks with fewer than 50 instances, where performance improvements exceed 10\% in some languages such as \textit{San20\_it}, \textit{Ous19\_ar}, and \textit{Xdomain\_tr}. While the performance gain decreases as more target language training data becomes available, the average results consistently show that leveraging cross-lingual data outperforms relying solely on the target language’s training set. 
In most languages—except for \textit{Bas19\_es} and \textit{Xdomain\_tr}—our proposed method outperforms the \texttt{Röttger} on average, while using less training data and without requiring manual selection of intermediate tasks. Notably, retrieving around 200 instances often yields comparable or even superior performance to this work, which uses 20{,}000 training size for intermediate fine-tuning.

% NEW
Another insight from Table~\ref{tab:evaluation_1} is how cross-lingual retrieval can compensate for limited labeled data in the target language. For example, in Hindi, retrieving just 20 instances for a training size of 20 matches the performance of having 50 labeled examples, and retrieving 2,000 instances approaches the performance of having 200 labeled instances. This pattern is consistent across other underlined values in the table.
In languages where \texttt{Mono} performance with 2{,}000 training samples fails to exceed 70\%—as in \textit{Ous19\_ar}, \textit{Ous19\_fr}, and \textit{Has21\_hi}—retrieval proves especially valuable, often matching the next training size. 

% NEW
For languages where \texttt{Mono}'s highest performance is less than 75\% (\textit{Gahd24\_de} and \textit{For19\_pt}), retrieval remains helpful, compensating for up to 500 labeled examples. However, in languages where \texttt{Mono} performance exceeds 75\% with 2,000 samples, retrieval is less beneficial—except in the extreme low-data case: with only 20 labeled data, retrieval consistently outperforms the \texttt{Mono} model trained on 50 examples across all languages.
% More detailed results and analysis are in Appendix~\ref{sec:full_results}.

\begin{figure*}[htbp]
  \centering

  \begin{subfigure}[b]{0.47\textwidth}
    \includegraphics[width=\textwidth]{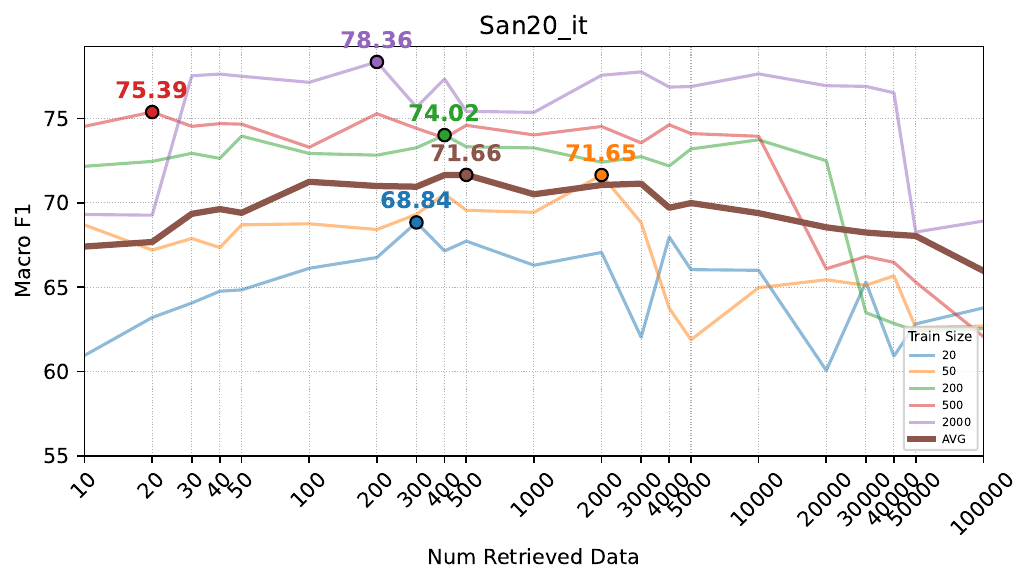}
  \end{subfigure}
  % \hfill
  \begin{subfigure}[b]{0.47\textwidth}
    \includegraphics[width=\textwidth]{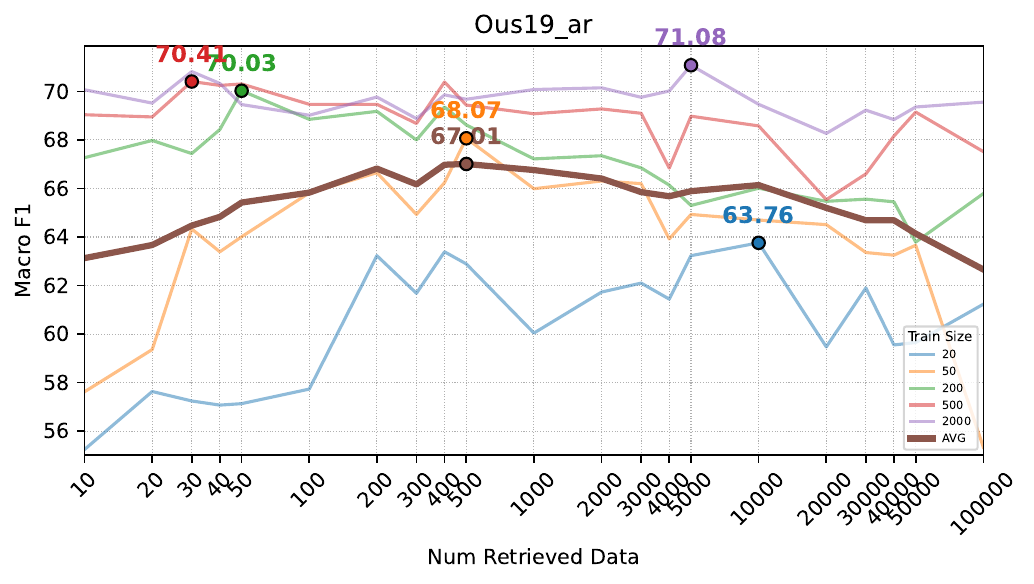}
  \end{subfigure}

  \begin{subfigure}[b]{0.47\textwidth}
    \includegraphics[width=\textwidth]{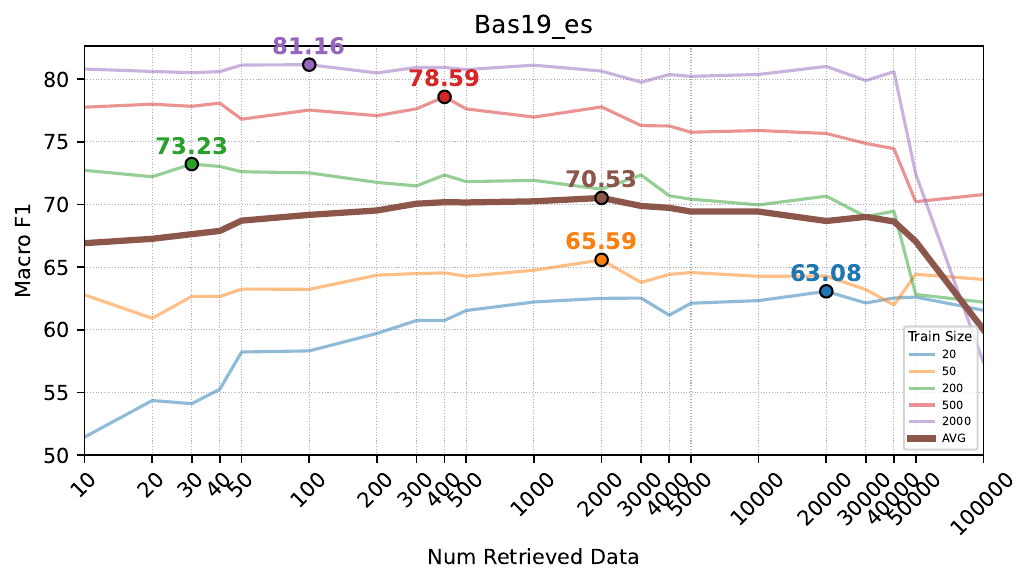}
  \end{subfigure}
  % \hfill
  \begin{subfigure}[b]{0.47\textwidth}
    \includegraphics[width=\textwidth]{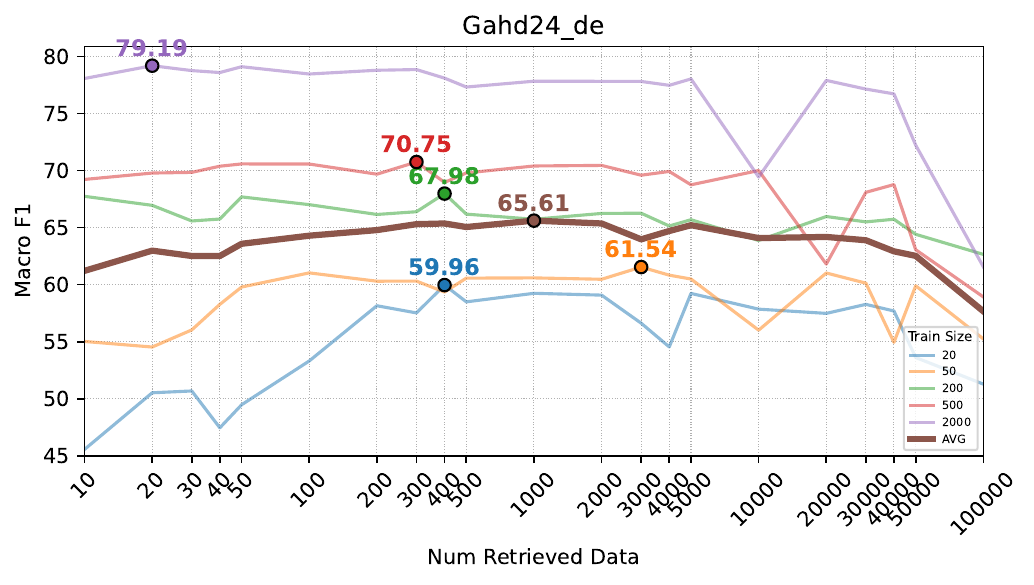}
  \end{subfigure}

  \caption{Performance across different numbers of retrieved instances (10 to 100{,}000, log-scaled) for four languages. Each curve corresponds to selected training sizes. The \texttt{AVG} line shows the average over 12 training sizes.}
  \label{fig:2x2grid}
\end{figure*}

\begin{figure}[t] % Force placement at the top
  \centering
  \includegraphics[width=0.95\linewidth]{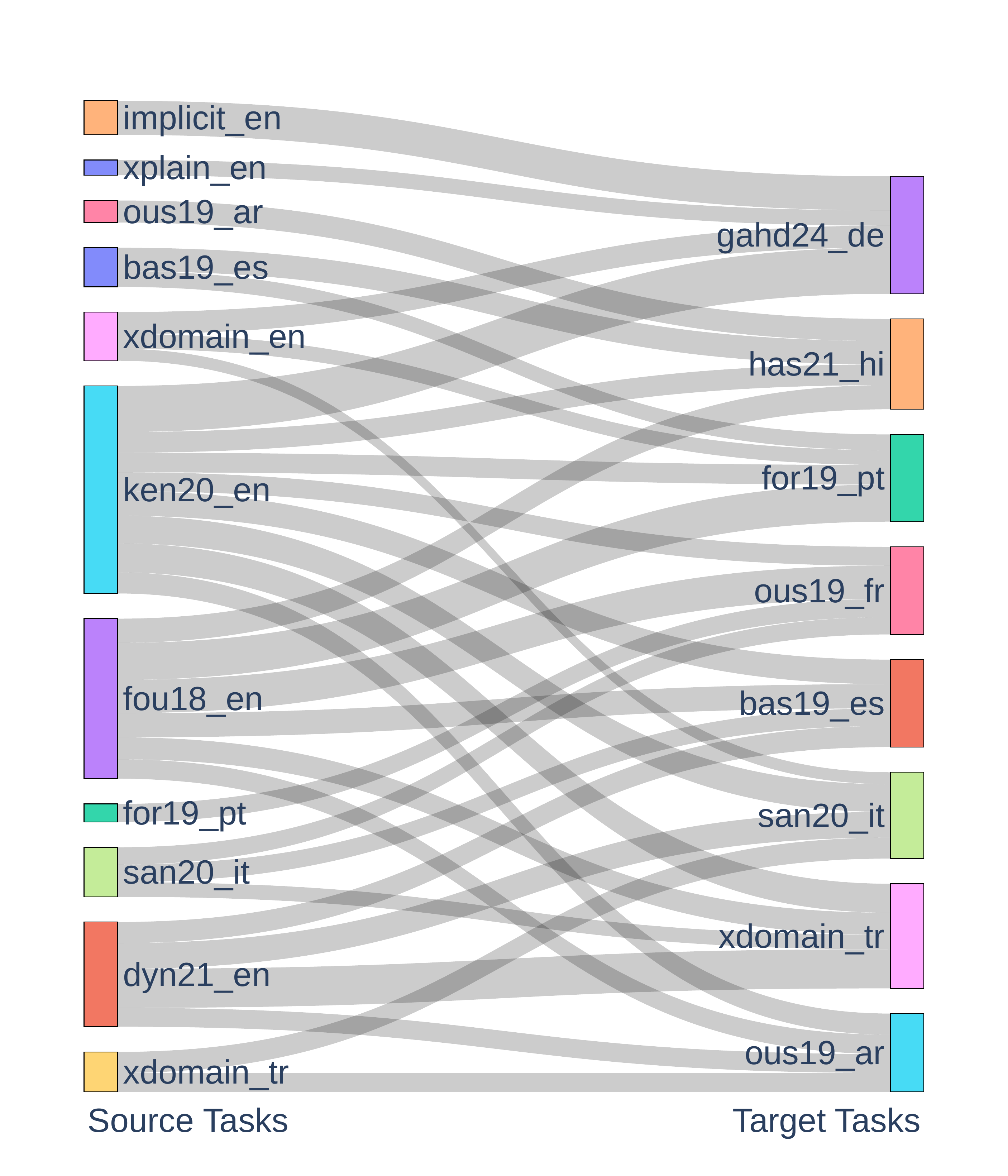} % Replace with your image file
  \caption{Sankey diagram of the distribution of the top four retrieved source tasks per target task.
  }
  \label{fig:sankey_tasks}
\end{figure}

Another observation from Table~\ref{tab:evaluation_1} is that, in five languages—excluding \textit{Ous19\_fr}, \textit{For19\_pt}, and \textit{Has21\_hi}—the highest average performance is achieved by retrieving 2{,}000 instances, while retrieving 20{,}000 leads to a performance drop. For instance, in \textit{Ous19\_ar}, retrieving 200 instances yields the best result. This suggests that increasing the number of retrieved data points for fine-tuning does not necessarily lead to improved performance.

%NEW
% \noindent
\paragraph{How Much Retrieved Data Is Sufficient?} To address this question, we conducted an experiment varying the number of retrieved instances across 21 settings, from 10 to 100{,}000 (More than a third of the pool size), for four languages as shown in Figure~\ref{fig:2x2grid}. The figure includes five different training sizes, each represented by a distinct color. The brown line labeled \texttt{AVG} denotes the average performance over 12 training sizes.

As shown in the figure, especially in the average trend line—where the effects of noise are diminished due to averaging— performance increases as the number of retrieved instances grows—up to around 2{,}000—after which it gradually declines. This change is more pronounced for smaller training sizes (e.g., 20), while for larger sizes (e.g., 2{,}000), the effect is minimal. These results suggest that adding more retrieved data is not always beneficial, and peak performance is typically reached with around 2{,}000 retrieved instances.

\subsection{Retrieved Languages Distribution}
An interesting analysis is to examine which tasks or languages the retrieved data come from for each target language. This is illustrated in Figure~\ref{fig:sankey_tasks}, which shows the average retrieval distribution when retrieving 2{,}000 instances for a training size of 2{,}000, averaged over five random subsamples of the original training set. In the Sankey diagram, source tasks are shown on the left and target tasks on the right, with edges representing the four most frequently retrieved source tasks for each target task. 
Due to the dominance of English data in the pool, a higher proportion of English instances is expected, with \textit{Ken20\_en} and \textit{Fou18\_en} being the most commonly retrieved source tasks.

However, we also observe non-negligible retrieval from smaller source tasks, such as Arabic, highlighting semantic and contextual relevance between hate speech in source and target languages.
%German primarily retrieved English sources, Portuguese used both English and Spanish, and Spanish relied on Italian and Turkish. Hindi retrieved from Spanish and Arabic, while Arabic used Turkish. 
%Turkish mainly drew from Italian, and French utilized Portuguese and Italian. 
% Overall, \textit{Ken20\_en} and \textit{Fou18\_en} are the most commonly retrieved source tasks, likely due to their size and generality.
We can also see that linguistically or culturally related languages tend to support each other: Portuguese benefits French, Turkish supports Arabic, and Italian aids Spanish. This highlights the effectiveness of our approach in identifying culturally proximate examples. This retrieval pattern can also be due to shared annotation styles or content overlap. Further diagrams are in Appendix \ref{sec:full_analysis}. 
% Although Turkish and German are the second- and third-largest languages in the pool, they are not retrieved as frequently as some other non-English tasks. In contrast, Spanish and Italian are retrieved more often. 
% This retrieval pattern may be due to regional or linguistic proximity as well as shared annotation styles, or content overlap. Further Sankey diagrams are in Appendix \ref{sec:full_analysis}. 

\begin{table*}[!htp]\centering
\scriptsize
\begin{tabular}{p{0.00001cm}p{0.01cm}p{0.5cm}p{0.4cm}p{0.4cm}p{0.4cm}p{0.4cm}p{0.00001cm}p{0.4cm}p{0.4cm}p{0.4cm}p{0.4cm}p{0.00001cm}p{0.4cm}p{0.4cm}p{0.4cm}p{0.4cm}p{0.00001cm}p{0.4cm}p{0.4cm}p{0.4cm}p{0.4cm}}\toprule
& &TRAIN&\multicolumn{4}{c}{{Bas19\_es}} & &\multicolumn{4}{c}{{For19\_pt}} & &\multicolumn{4}{c}{{Has21\_hi}} & &\multicolumn{4}{c}{{Ous19\_fr}} \\\cmidrule{4-7}\cmidrule{9-12}\cmidrule{14-17}\cmidrule{19-22}
& &SIZE &20 &200 &2{,}000 &20{,}000 & &20 &200 &2{,}000 &20{,}000 & &20 &200 &2{,}000 &20{,}000 & &20 &200 &2{,}000 &20{,}000 \\\midrule
\multirow{5}{*}{\rotatebox{90}{English}} &\multirow{5}{*}{\rotatebox{90}{Retrieval}} &20 &\textbf{55.83} &\textbf{61.15} &61.92 &63.03 & &\textbf{52.00} &\textbf{65.73} &68.11 &67.55 & &46.92 &\textbf{52.72} &\textbf{56.01} &\textbf{55.71} & &47.26 &51.17 &53.58 &55.10 \\
& &50 &\textbf{61.67} &\textbf{65.19} &64.13 &62.91 & &57.58 &66.86 &\textbf{67.71} &68.52 & &48.90 &52.67 &\textbf{55.62} &55.46 & &\textbf{48.64} &51.56 &54.38 &53.35 \\
& &200 &71.53 &71.08 &71.01 &68.24 & &69.13 &70.04 &68.73 &70.19 & &54.22 &\textbf{56.24} &\textbf{58.62} &56.44 & &50.70 &\textbf{56.15} &54.11 &53.56 \\
& &500 &77.13 &76.28 &76.67 &74.55 & &69.60 &70.67 &\textbf{70.77} &70.02 & &55.23 &57.54 &\textbf{59.83} &58.02 & &51.75 &53.22 &52.46 &53.48 \\
& &2{,}000 &80.67 &80.68 &80.68 &80.54 & &72.07 &72.32 &71.12 &64.82 & &57.40 &59.00 &\textbf{60.63} &\textbf{60.74} & &50.47 &53.11 &52.31 &53.82 \\\cmidrule{4-7}\cmidrule{9-12}\cmidrule{14-17}\cmidrule{19-22}
& &AVG &67.27 &\textbf{69.98} &69.61 &68.62 & &62.86 &\textbf{68.53} &69.02 &68.32 & &52.04 &\textbf{55.22} &\textbf{57.19} &56.24 & &50.00 &52.40 &53.38 &54.07 \\\midrule
\multirow{5}{*}{\rotatebox{90}{Multilingual}} &\multirow{5}{*}{\rotatebox{90}{Retrieval}} &20 &54.37 &59.72 &\textbf{62.52} &63.08 & &49.72 &64.92 &68.57 &\textbf{68.03} & &\textbf{47.34} &51.03 &53.68 &55.37 & &47.21 &\textbf{52.68} &53.93 &55.05 \\
& &50 &60.93 &64.37 &\textbf{65.59} &\textbf{64.30} & &59.26 &\textbf{67.01} &67.06 &\textbf{69.35} & &\textbf{48.39} &\textbf{53.36} &52.26 &\textbf{55.78} & &48.29 &52.19 &52.97 &\textbf{55.60} \\
& &200 &\textbf{72.22} &71.77 &71.23 &70.67 & &\textbf{69.69} &\textbf{70.33} &\textbf{70.20} &\textbf{71.07} & &\textbf{55.83} &54.65 &56.80 &56.02 & &51.54 &54.06 &\textbf{55.80} &53.63 \\
& &500 &\textbf{78.01} &77.09 &\textbf{77.79} &67.67 & &69.72 &\textbf{70.84} &70.04 &\textbf{71.05} & &\textbf{56.94} &57.66 &57.88 &\textbf{59.55} & &\textbf{53.30} &52.84 &\textbf{55.51} &\textbf{55.31} \\
& &2{,}000 &80.62 &80.50 &80.65 &\textbf{81.02} & &\textbf{72.39} &\textbf{72.66} &71.72 &\textbf{72.22} & &\textbf{58.19} &\textbf{60.22} &60.50 &59.65 & &\textbf{53.51} &53.13 &\textbf{53.30} &\textbf{54.74} \\\cmidrule{4-7}\cmidrule{9-12}\cmidrule{14-17}\cmidrule{19-22}
& &AVG &67.27 &69.53 &\textbf{70.53} &68.69 & &62.85 &68.18 &\textbf{69.55} &\textbf{69.69} & &\textbf{52.44} &55.10 &56.25 &\textbf{57.05} & &5\textbf{0.56} &\textbf{53.12} &\textbf{54.05} &\textbf{54.84} \\
\bottomrule
\end{tabular}
\caption{F1-macro scores across two strategies: English-only retrieval, and Multilingual retrieval. 
%Results are shown for target training sizes of 20, 50, 200, 500, 2{,}000, and \texttt{AVG} (the average over 12 training sizes). Columns represent target languages, and sub-columns are the number of retrieved instances.
}\label{tab:ablation_results}
\end{table*}

\subsection{English-only vs Multilingual Retrieval}
This experiment examines the effect of multilingual retrieval by comparing it to English-only retrieval, where data is retrieved exclusively from English tasks. Table~\ref{tab:ablation_results} presents the results: rows are retrieval settings, and columns represent four target languages (see Appendix~\ref{sec:english_vs_multilingual} for other languages). Comparisons should be made vertically within each language—for example, comparing 20 training samples with 20 retrieved instances across the two row blocks.
We observe only minor differences in overall performance across the two settings in the table, likely due to the high proportion of English data in the pool. However, in specific cases—such as retrieving 2{,}000 instances for \textit{Bas19\_es} and \textit{For19\_pt}, and 20 or 2{,}000 instances for \textit{Has21\_hi} and \textit{Ous19\_fr}—multilingual retrieval yields higher performance. This suggests that incorporating even a small amount of multilingual data can be beneficial.

\section{Maximum Marginal Relevance}
As an additional deduplication step, we apply \emph{Maximum Marginal Relevance (MMR)} in the retrieval module—before mapping the retrieved vectors back to their original texts—to ensure both relevance and diversity. Specifically, we retrieve at least \( 2R \) candidate vectors and iteratively select \( R \) vectors that balance similarity to the query and dissimilarity to previously selected vectors.
Given a query vector \( q \), a candidate set \( D \), and a selected set \( S \), MMR selects the next vector \( v^* \in D \setminus S \) as:

\begin{multline*}
\text{MMR}(v^*) = \arg\max_{v \in D \setminus S} \Big[ 
\lambda \cdot \cos(v, q) \\
- (1 - \lambda) \cdot \max_{s \in S} \cos(v, s) \Big]
\end{multline*}

\noindent
Here, \( \lambda \in [0, 1] \) controls the trade-off between relevance to the query and diversity with respect to the selected set. We set the $\lambda=0.5$. This process is repeated until exactly \( R \) vectors are selected.

Although removing highly similar instances using MMR increases the diversity of the retrieved data, however, incorporating it does not affect performance a lot, with results remaining largely similar across most languages—except for those listed in Table~\ref{tab:mmr} (See Appendix \ref{sec:full_mmr}). As shown in the figure, for these three languages, applying MMR particularly improves performance when retrieving fewer than 2{,}000 instances. In contrast, for 20{,}000 retrieved instances, the performance without MMR is higher. 
%This is because MMR involves retrieving twice as many candidates (2\textit{k}) and then filtering out the most similar ones to retain \textit{k} diverse instances.
This suggests that when only a limited number of instances is retrieved, MMR helps select fewer but more diverse examples, which can lead to improved performance. Interestingly, for the Turkish dataset—where our method previously underperformed without MMR—applying it allows the model to surpass the performance of \texttt{Röttger}.

\begin{table}[!htp]\centering
\scriptsize
\begin{tabular}{p{0.01cm}p{0.3cm}p{0.35cm}p{0.35cm}p{0.35cm}p{0.35cm}p{0.0000001cm}p{0.35cm}p{0.35cm}p{0.35cm}p{0.35cm}}\toprule
&TRAIN &\multicolumn{4}{c}{Without MMR} & &\multicolumn{4}{c}{With MMR} \\\cmidrule{3-6}\cmidrule{8-11}
&SIZE &20 &200 &2{,}000 &20{,}000 & &20 &200 &2{,}000 &20{,}000 \\\midrule
\multirow{5}{*}{\rotatebox{90}{San20\_it}} &20 &\textbf{63.20} &66.76 &\textbf{67.06} &60.06 & &60.28 &\textbf{66.89} &61.69 &\textbf{62.38} \\
&50 &67.20 &\textbf{68.42} &\textbf{71.65} &\textbf{69.44} & &\textbf{69.71} &68.30 &70.11 &63.30 \\
&200 &72.46 &\textbf{72.83} &72.41 &\textbf{72.50} & &72.51 &72.66 &\textbf{73.07} &72.08 \\
&500 &\textbf{75.39} &75.29 &74.53 &66.09 & &74.93 &\textbf{75.41} &\textbf{75.00} &\textbf{75.69} \\
&2{,}000 &69.27 &\textbf{78.36} &77.57 &\textbf{76.95} & &\textbf{77.66} &77.06 &\textbf{76.67} &68.88 \\\cmidrule{3-6}\cmidrule{8-11}
&AVG &67.68 &71.00 &71.06 &\textbf{68.55} & &\textbf{69.24} &\textbf{71.64} &\textbf{71.38} &68.33 \\\midrule
\multirow{5}{*}{\rotatebox{90}{Gahd24\_de}} &20 &50.52 &58.15 &\textbf{59.08} &57.48 & &\textbf{51.47} &58.14 &58.63 &\textbf{59.14} \\
&50 &54.53 &60.30 &60.47 &\textbf{61.02} & &54.83 &60.07 &\textbf{61.13} &60.50 \\
&200 &66.95 &66.15 &66.24 &\textbf{65.97} & &\textbf{68.09} &66.20 &\textbf{67.13} &65.41 \\
&500 &69.78 &69.68 &70.45 &61.80 & &\textbf{70.36} &\textbf{70.11} &70.06 &\textbf{62.77} \\
&2{,}000 &\textbf{79.19} &78.79 &77.82 &\textbf{77.90} & &78.55 &\textbf{79.08} &77.53 &68.84 \\\cmidrule{3-6}\cmidrule{8-11}
&AVG &62.98 &64.78 &65.36 &\textbf{64.18} & &62.71 &\textbf{65.13} &\textbf{65.83} &63.62 \\\midrule
\multirow{5}{*}{\rotatebox{90}{Xdomain\_tr}} &20 &\textbf{66.58} &67.08 &70.14 &75.87 & &65.19 &\textbf{72.00} &\textbf{77.16} &\textbf{67.48} \\
&50 &75.92 &\textbf{77.50} &\textbf{78.85} &70.60 & &75.98 &76.08 &\textbf{79.84} &64.58 \\
&200 &81.61 &\textbf{82.61} &83.04 &\textbf{82.61} & &81.43 &80.92 &83.06 &82.76 \\
&500 &84.93 &\textbf{85.09} &\textbf{84.92} &\textbf{83.88} & &84.77 &84.60 &84.01 &83.41 \\
&2{,}000 &87.39 &\textbf{88.00} &\textbf{87.39} &77.48 & &\textbf{88.14} &87.73 &86.91 &66.70 \\\cmidrule{3-6}\cmidrule{8-11}
&AVG &76.78 &78.88 &79.66 &\textbf{77.58} & &\textbf{77.59} &\textbf{79.75} &\textbf{80.80} &76.01 \\
\bottomrule
\end{tabular}
\caption{F1-macro scores without/with MMR for three languages (rows).
%across five selected training sizes and an average (\texttt{AVG}) computed over 12 training sizes.
}\label{tab:mmr}
\end{table}

\section{Conclusion}
This paper presents a cross-lingual nearest neighbor retrieval approach to improve hate speech detection in target languages with limited labeled data. Our method retrieves the nearest neighbors from a multilingual pool of source tasks to augment the target language data, consistently outperforming models trained solely on the target language. Notably, even with as few as 20 labeled instances in the target language, our approach can achieve performance improvements of up to 10\% in some cases. Further, we show that retrieving approximately 2,000 instances yields the highest average performance, while retrieving more can lead to a performance drop. 
%Additionally, applying MMR to filter out redundant data can further improve results in certain settings. 
Our method is scalable and adaptable to new languages and tasks, 
%as it is easy to integrate and extend, 
allowing new source tasks to be added to the pool with minimal effort.

\section*{Limitations}
Despite the effectiveness of our approach, several limitations remain. First, we assume access to a small number of labeled hate speech instances in the target language. While this assumption reduces annotation cost, it may not hold in extremely low-resource settings where even minimal labeled data is unavailable or difficult to obtain due to linguistic, political, or ethical constraints. 

Second, the retrieval pool used in our experiments is heavily imbalanced, with English accounting for the majority of instances. This dominance can bias retrieval and limit performance improvements for target languages that are typologically distant or culturally distinct from English. 

While we reviewed the definitions of hate speech used in the datasets for our experiments, cultural differences and annotation inconsistencies may still be present. Although hate speech is undoubtedly influenced by cultural context, many hateful expressions are universal across languages and cultures. Our experiments demonstrate that leveraging such cross-lingual data can effectively improve hate speech detection in low-resource settings.

Our evaluation focuses on a subset of hate speech detection tasks and languages and does not encompass the full variety of online abuse domains or contexts in which hate speech occurs. Expanding the set of target labels and tasks, especially in non-Western languages and underrepresented communities, would help assess the robustness and generalizability of the proposed method.

Finally, while our results confirm the effectiveness of cross-lingual retrieval for hate speech detection, a promising direction for future work is to more deeply analyze the nature and influence of the retrieved instances. Such analysis could offer further insights into why retrieval is effective, enhance our understanding of cross-lingual transfer, similarities, and differences of hate speech across languages, and contribute to more interpretable and adaptive retrieval strategies.

\section*{Ethical Consideration}
Our approach relies on the availability of existing labeled datasets for hate speech detection, many of which are restricted to non-commercial research use. This presents limitations for real-world deployment, particularly in industry settings where such licensing terms may prohibit direct reuse. However, the proposed method remains valuable in industrial scenarios where systems are initially developed for high-resource languages and later adapted to lower-resource markets. Instead of incurring substantial costs to collect large-scale labeled data from scratch, companies could use this cross-lingual transfer approach to quickly build minimum viable products (MVPs) for new linguistic contexts.

Still, while our method provides an efficient path for adapting models to low-resource languages using cross-lingual retrieval, its applicability is ultimately constrained by legal access to source data. Therefore, we emphasize that the responsible use of our method and, further, data for the downstream cases requires careful attention to data licensing and ethical data sharing practices.

%Another limitation is that our approach relies on the assumption that nearest neighbors in the embedding space are semantically useful across tasks and languages. This may not always hold true, especially for low-resource or underrepresented languages where the embedding model may not capture subtle linguistic or cultural nuances. Moreover, we do not explicitly address fairness or annotation bias in the datasets, which can affect both retrieval and downstream classification quality. 

% \section*{Ethics Statement}

% \section*{Acknowledgements}

\section*{Acknowledgements}
The work was supported by the European Research Council (ERC) through the European Union's Horizon Europe research and innovation programme (grant agreement No. 101113091) and the German Research Foundation (DFG; grant FR 2829/7-1).

\bibliography{custom}
\bibliographystyle{acl_natbib}

\onecolumn
\appendix

\section{Datasets Details}
\label{sec:datasets_details}
We used fourteen datasets in our study. Detailed information—including language, number of instances, license type—is provided in Table~\ref{tab:detailed_dataset_information}. In total, we had 265,671 instances, of which 62.85\% were Non-Hate and 37.15\% were Hate Speech.

\begin{table}[!htp]\centering
\scriptsize
\begin{tabular}{llp{0.9cm}p{1cm}p{1cm}p{1cm}p{1cm}p{3cm}}\toprule
Name &Language &Language Code &Num of Instances & \% of Total Instances &Num classes &\% of Hate Speech &Licence\\\midrule
Bas19\_es~\citep{basile-etal-2019-semeval} &Spanish &es &6600 &2.48 &2 &41.50 & \hyperlink{https://creativecommons.org/licenses/by/4.0/}{CC BY 4.0}\\
For19\_pt~\citep{fortuna-etal-2019-hierarchically} &Purtegues &pt &5670 &2.13&2 &31.53 & \hyperlink{https://creativecommons.org/licenses/by/4.0/}{CC BY 4.0}\\
Has21\_hi~\citep{mandl2021overview} &Hindi &hi &4594 &1.73 &2 &12.32 & \hyperlink{https://creativecommons.org/licenses/by/4.0/}{CC BY 4.0} (Only for research purposes.)\\
Ous19\_ar~\citep{ousidhoum-etal-2019-multilingual} &Arabic &ar &3353 &1.26&2 &22.52 & MIT\\
Ous19\_fr~\citep{ousidhoum-etal-2019-multilingual} &French &fr &4014 &1.51&2 &00.94 & MIT \\
San20\_it~\citep{sanguinetti2020haspeede} &Italian &it &8100 &3.05&2 &41.83 & \hyperlink{https://creativecommons.org/licenses/by-nc-sa/4.0/deed.en}{CC BY-NC-SA 4.0}\\
Gahd24\_de~\citep{goldzycher-etal-2024-improving} &German &de &10996 &3.84&2 &42.37 & \hyperlink{https://creativecommons.org/licenses/by/4.0/}{CC BY 4.0}\\\
Xdomain\_tr~\citep{toraman-etal-2022-large}&Turkish &tr &37933 &17.0&2 &42.67 & \hyperlink{https://creativecommons.org/licenses/by-nc-sa/4.0/deed.en}{CC BY-NC-SA 4.0}\\
Ken20\_en~\citep{kennedy2020contextualizing} &English &en &23192 &8.73&2 &50.00 & MIT\\
Fou18\_en~\citep{founta2018large} &English &en &22565 &8.49&2 &22.00 & \hyperlink{https://creativecommons.org/licenses/by/4.0/}{CC BY 4.0}\\
Xplain\_en~\citep{mathew2021hatexplain} &English &en &13749 &5.08&2 &43.22 & MIT\\
Implicit\_en~\citep{elsherief-etal-2021-latent} &English &en &21480 &8.09&2 &38.12 & MIT\\
Dyn21\_en~\citep{vidgen-etal-2021-learning} &English &en &41144 &15.49&2 &46.10 & \hyperlink{https://creativecommons.org/licenses/by/4.0/}{CC BY 4.0}\\
Xdomain\_en~\citep{toraman-etal-2022-large}&English &en &47124 &21.12&2 &19.41 & \hyperlink{https://creativecommons.org/licenses/by-nc-sa/4.0/deed.en}{CC BY-NC-SA 4.0}\\
\bottomrule
\end{tabular}
\caption{Detailed information about the datasets used in this study.}\label{tab:detailed_dataset_information}
\end{table}

\section{Model and Training Details}
\label{sec:models_details}
\subsection{Embedder}
For the embedding model, we used \texttt{BAAI/bge-m3},\footnote{\url{https://huggingface.co/BAAI/bge-m3}}$^{,}$\footnote{\url{https://github.com/FlagOpen/FlagEmbedding}} accessed via the Sentence Transformers library.\footnote{\url{https://github.com/UKPLab/sentence-transformers}} This model supports over 100 languages, is effective for both short and long text retrieval, and produces 1024-dimensional embeddings. It is released under the MIT license, and the Sentence Transformers library is licensed under Apache 2.0—both allowing use in academic research. We used the model in inference mode without any fine-tuning, applying it to our text data to generate embedding vectors.
\subsection{Retriever}
For indexing and searching the embedding vectors in the retrieval pool, we used the \texttt{Faiss}, library\footnote{\url{https://github.com/facebookresearch/faiss}} which is licensed under MIT. We employed the HNSW (Hierarchical Navigable Small World) index with Euclidean distance as the similarity metric, where smaller values indicate greater similarity to the query. Since the size of the retrieval pool was moderate, we used the CPU version of the library. The index was configured with 128 neighbors, a construction parameter of 200, and a search parameter of 128.

\subsection{Fine-tuner}
We used \texttt{cardiffnlp/twitter-xlm-roberta-base}\footnote{\url{https://huggingface.co/cardiffnlp/twitter-xlm-roberta-base}} (XLM-T), a multilingual transformer-based model pre-trained on Twitter data, as our base model for fine-tuning on hate speech detection tasks. The model is licensed under Apache 2.0, which permits use in academic research. Training and evaluation were performed using the Hugging Face \texttt{transformers} library.\footnote{\url{https://github.com/huggingface/transformers}}

To select the classification model, we also evaluated \texttt{xlm-roberta-base}\footnote{\url{https://huggingface.co/FacebookAI/xlm-roberta-base}} (XLM-R) and \texttt{mdeberta-v3-base}\footnote{\url{https://huggingface.co/microsoft/mdeberta-v3-base}} \citep{he2021debertav3}. XLM-T outperformed XLM-R, likely because hate speech datasets are primarily sourced from social media, where XLM-T has been pre-trained. Although XLM-T and mDeBERTa achieved similar performance, we chose to use XLM-T in our experiments, as it supports a broader range of languages and aligns with our baseline setting.

Throughout all our experiments, for training sizes with fewer than 9{,}999 training instances, we trained for 10 epochs; for larger training sizes, we used 5 epochs to reduce training time and avoid overfitting. We set the batch size to 16 and used a learning rate of 5e-5. Inputs were truncated or padded to a maximum sequence length of 128 tokens. We used binary cross-entropy loss, as our datasets involved binary classification (Hate vs. Non-Hate). All other training hyperparameters were left at their default values provided by the \texttt{transformers.Trainer} module.

\section{Hardware, Tools, and Ethical Compliance}
The experiments were conducted on NVIDIA GeForce GTX 1080 Ti servers. The embedding model was used in inference mode without updating its parameters, while the classification model was fully fine-tuned. As the classifier was based on \texttt{xlm-roberta-base}, it included approximately 279 million parameters.
We also acknowledge the use of an AI assistant during the writing process. ChatGPT\footnote{\url{https://chatgpt.com/}} was used for paraphrasing and improving clarity throughout the formulation of the paper. All models and datasets used in this study are licensed for academic research purposes and align with the intended use of advancing NLP applications for social good.

%NEW
\section{Further Results}
\label{sec:full_results}
Table~\ref{tab:full_main_table} presents the full results of cross-lingual nearest neighbor retrieval fine-tuning, covering training sizes from 10 to 2{,}000. These results confirm the trend shown in the main table (Table~\ref{tab:evaluation_1}): retrieving as few as 20 instances already outperforms the \texttt{Mono} baseline in all languages. These performance improvements are most notable when the target language has fewer than 50 training examples, with gains exceeding 10\% compared to training solely on the target language data. In such low-resource settings, retrieving cross-lingual nearest neighbors and using them for fine-tuning enables the model to match the performance of much larger training sizes.
 In most cases, our method also outperforms the \texttt{Röttger} approach, while using less data and without requiring manual selection of a source task.

\begin{table}[!htp]\centering
\scriptsize
\begin{tabular}{p{0.5cm}p{0.35cm}p{0.35cm}p{0.35cm}p{0.35cm}p{0.35cm}p{0.45cm}p{0.00001cm}p{0.35cm}p{0.35cm}p{0.35cm}p{0.35cm}p{0.35cm}p{0.45cm}p{0.00001cm}p{0.35cm}p{0.35cm}p{0.35cm}p{0.35cm}p{0.35cm}p{0.45cm}}\toprule
&\multicolumn{6}{c}{San20\_it} & &\multicolumn{6}{c}{Ous19\_ar} & &\multicolumn{6}{c}{Ous19\_fr} \\\cmidrule{2-7}\cmidrule{9-14}\cmidrule{16-21}
SIZE &Mono &20 &200 &2000 &20000 &Röttger & &Mono &20 &200 &2000 &20000 &Röttger & &Mono &20 &200 &2000 &20000 &Röttger \\\midrule
10 &46.23 &47.04 &\underline{64.96} &\textbf{65.26} &69.60 &63.34 & &51.98 &\underline{54.60} &56.36 &\textbf{62.86} &61.54 &59.00 & &47.26 &48.34 &\underline{52.46} &51.90 &\textbf{54.60} &53.49 \\
20 &54.25 &\underline{63.20} &66.76 &\textbf{67.06} &60.06 &64.96 & &51.67 &\underline{57.63} &\textbf{63.23} &61.73 &59.47 &60.52 & &47.26 &47.21 &\underline{52.68} &53.93 &\textbf{55.05} &52.93 \\
30 &56.29 &\underline{64.24} &68.47 &68.90 &\textbf{70.08} &64.95 & &44.42 &\underline{57.95} &\textbf{66.23} &62.31 &62.70 &65.48 & &47.26 &47.60 &\underline{52.82} &53.86 &\textbf{54.97} &53.91 \\
40 &59.14 &59.60 &\underline{67.91} &69.16 &\textbf{70.58} &67.41 & &49.67 &\underline{57.45} &\textbf{65.70} &64.52 &64.82 &64.29 & &47.24 &48.36 &\underline{52.40} &54.63 &56.30 &\textbf{56.62} \\
50 &65.71 &67.20 &68.42 &\underline{71.65} &\textbf{69.44} &69.10 & &52.13 &59.36 &\underline{\textbf{66.65}} &66.31 &64.51 &65.76 & &47.87 &48.29 &\underline{52.19} &52.97 &\textbf{55.60} &54.15 \\
100 &70.96 &70.01 &71.17 &71.46 &67.73 &\textbf{71.89} & &65.29 &66.22 &\underline{\textbf{68.50}} &66.77 &66.07 &66.44 & &47.67 &49.35 &\underline{51.54} &53.12 &\textbf{56.04} &55.83 \\
200 &72.81 &72.46 &\underline{\textbf{72.83}} &72.41 &72.50 &71.56 & &67.97 &\underline{67.98} &\textbf{69.18} &67.35 &65.47 &66.61 & &51.93 &\underline{51.54} &54.06 &\textbf{55.80} &53.63 &53.76 \\
300 &73.43 &73.56 &\underline{\textbf{74.12}} &72.03 &64.56 &72.21 & &66.95 &\underline{68.94}&68.52 &\textbf{69.00} &66.71 &68.07 & &51.10 &\underline{53.49} &53.29 &54.58 &\textbf{56.15} &53.61 \\
400 &72.44 &73.49 &\underline{\textbf{74.84}} &66.23 &66.80 &73.32 & &67.04 &\underline{69.75} &\textbf{70.20} &68.88 &68.20 &66.79 & &52.17 &\underline{53.22} &54.57 &53.81 &\textbf{55.13} &53.34 \\
500 &74.18 &75.39 &\textbf{75.29} &74.53 &66.09 &73.69 & &66.54 &\underline{68.95} &\textbf{69.47} &69.28 &65.54 &67.60 & &51.91 &\underline{53.30} &52.84 &\textbf{55.51} &55.31 &53.39 \\
1000 &76.56 &\underline{\textbf{76.65}} &68.84 &76.41 &68.21 &76.14 & &67.29 &\underline{68.98} &68.08 &67.77 &\textbf{69.07} &67.26 & &53.14 &\underline{52.49} &\textbf{55.48} &55.15 &50.51 &52.59 \\
2000 &76.40 &69.27 &\textbf{78.36} &77.57 &76.95 &77.07 & &66.91 &69.52 &69.77 &\textbf{70.15} &68.27 &67.07 & &51.84 &53.51 &53.13 &53.30 &\textbf{54.74} &52.89 \\\cmidrule{2-7}\cmidrule{9-14}\cmidrule{16-21}
AVG &66.53 &67.68 &71.00 &\textbf{71.06} &68.55 &70.47 & &59.82 &63.94 &\textbf{66.82} &66.41 &65.20 &65.41 & &49.72 &50.56 &53.12 &54.05 &\textbf{54.84} &53.88 \\\midrule
&\multicolumn{6}{c}{Bas19\_es} & &\multicolumn{6}{c}{For19\_pt} & &\multicolumn{6}{c}{Xdomain\_tr} \\\cmidrule{2-7}\cmidrule{9-14}\cmidrule{16-21}
10 &36.51 &46.06 &\underline{58.18} &\textbf{62.23} &61.97 &59.71 & &43.18 &48.03 &\underline{61.67} &67.53 &\textbf{67.79} &66.38 & &46.92 &\underline{56.19} &73.20 &\textbf{77.11} &72.50 &72.50 \\
20 &49.91 &54.37 &\underline{59.72} &62.52 &63.08 &\textbf{66.52} & &48.09 &49.72 &\underline{64.92} &\textbf{68.57} &68.03 &67.68 & &55.43 &\underline{66.58} &67.08 &70.14 &\textbf{75.87} &69.80 \\
30 &54.38 &\underline{60.75} &62.56 &64.62 &63.52 &\textbf{69.02} & &51.73 &\underline{54.34} &64.61 &\textbf{69.70} &67.80 &67.78 & &58.90 &\underline{73.13} &73.64 &\textbf{78.37} &75.58 &75.58 \\
40 &57.63 &57.69 &\underline{62.77} &63.10 &61.87 &\textbf{66.98} & &53.83 &52.63 &\underline{65.59} &\textbf{69.68} &67.57 &67.04 & &69.97 &\underline{74.20} &75.43 &61.71 &\textbf{78.70} &75.98 \\
50 &61.85 &60.93 &64.37 &\underline{65.59} &64.30 &\textbf{70.36} & &60.25 &59.26 &\underline{67.01} &67.06 &\textbf{69.35} &66.51 & &72.24 &\underline{75.92} &77.50 &\textbf{78.85} &70.60 &75.12 \\
100 &65.03 &65.36 &66.04 &67.71 &65.93 &\textbf{71.94} & &64.38 &\underline{67.81} &68.26 &\textbf{69.41} &68.99 &68.95 & &71.43 &79.84 &78.77 &80.79 &80.48 &\textbf{81.41} \\
200 &72.36 &72.22 &71.77 &71.23 &70.67 &\textbf{75.27} & &66.91 &\underline{69.69} &70.33 &70.20 &\textbf{71.07} &68.10 & &81.63 &81.61 &\underline{82.61} &\textbf{83.04} &82.61 &82.19 \\
300 &74.40 &76.04 &76.18 &75.31 &71.92 &\textbf{76.43} & &68.86 &\underline{69.37} &69.97 &69.86 &\textbf{70.14} &68.63 & &81.34 &81.80 &83.74 &83.27 &83.39 &\textbf{84.36} \\
400 &76.58 &75.77 &76.06 &76.30 &73.84 &\textbf{77.57} & &69.10 &69.74 &\underline{70.19} &69.96 &\textbf{70.80} &67.92 & &84.54 &83.08 &84.53 &84.27 &63.18 &\textbf{85.24} \\
500 &77.14 &78.01 &77.09 &77.79 &67.67 &\textbf{78.76} & &69.95 &69.72 &70.84 &70.04 &\underline{\textbf{71.05}} &69.22 & &85.05 &84.93 &85.09 &84.92 &83.88 &\textbf{85.34} \\
1000 &79.35 &79.42 &79.16 &79.27 &78.53 &\textbf{81.06} & &70.97 &71.48 &\textbf{72.07} &70.81 &71.44 &70.92 & &87.01 &76.68 &77.00 &86.03 &86.67 &\textbf{86.86} \\
2000 &81.08 &80.62 &80.50 &80.65 &81.02 &\textbf{82.04} & &72.70 &72.39 &\textbf{72.66} &71.72 &72.22 &71.61 & &88.53 &87.39 &88.00 &87.39 &77.48 &\textbf{88.84} \\\cmidrule{2-7}\cmidrule{9-14}\cmidrule{16-21}
AVG &65.52 &67.27 &69.53 &70.53 &68.69 &\textbf{72.97} & &61.66 &62.85 &68.18 &69.55 &\textbf{69.69} &68.39 & &73.58 &76.78 &78.88 &79.66 &77.58 &\textbf{80.27} \\\midrule
&\multicolumn{6}{c}{Gahd24\_de} & &\multicolumn{6}{c}{Has21\_hi} & & & & & & & \\\cmidrule{2-7}\cmidrule{9-14}
10 &38.03 &\underline{51.17} &54.50 &\textbf{59.85} &57.59 &58.88 & &46.87 &49.00 &\underline{51.76} &53.06 &\textbf{56.24} &54.46 & & & & & & & \\
20 &44.99 &\underline{50.52} &58.15 &59.08 &57.48 &\textbf{59.82} & &46.87 &47.34 &\underline{51.03} &53.68 &\textbf{55.37} &54.92 & & & & & & & \\
30 &50.20 &53.14 &\underline{59.37} &\textbf{60.57} &59.08 &59.78 & &46.87 &46.87 &\underline{52.31} &54.21 &55.46 &\textbf{57.47} & & & & & & & \\
40 &57.47 &55.18 &\underline{58.25} &\textbf{60.86} &57.79 &60.75 & &46.87 &47.67 &\underline{53.86} &55.58 &\textbf{56.33} &54.08 & & & & & & & \\
50 &57.85 &54.53 &60.30 &60.47 &61.02 &\textbf{62.57} & &46.87 &48.39 &\underline{53.36} &52.26 &\textbf{55.78} &54.77 & & & & & & & \\
100 &62.23 &64.27 &63.29 &62.93 &62.58 &\textbf{64.50} & &48.94 &51.38 &\underline{54.90 }&55.53 &56.91 &\textbf{57.88} & & & & & & & \\
200 &65.80 &\underline{66.95} &66.15 &\textbf{66.24} &65.97 &64.25 & &52.20 &\underline{55.83} &54.65 &56.80 &56.02 &\textbf{57.47} & & & & & & & \\
300 &67.17 &\underline{67.81} &\textbf{67.33} &64.70 &66.97 &64.58 & &51.75 &\underline{55.80} &55.43 &57.24 &\textbf{58.05} &57.51 & & & & & & & \\
400 &66.82 &\underline{69.04} &68.47 &68.43 &\textbf{69.52} &66.74 & &54.77 &54.50 &55.98 &\underline{\textbf{58.50} }&57.06 &58.12 & & & & & & & \\
500 &66.56 &69.78 &69.68 &\underline{\textbf{70.45}} &61.80 &67.02 & &56.20 &\underline{56.94} &57.66 &57.88 &\textbf{59.55} &57.96 & & & & & & & \\
1000 &69.81 &\underline{\textbf{74.17}} &73.02 &72.92 &72.50 &69.45 & &56.18 &\underline{57.40} &\textbf{60.06} &59.76 &58.14 &57.74 & & & & & & & \\
2000 &73.77 &\textbf{79.19} &78.79 &77.82 &77.90 &72.42 & &57.14 &58.19 &60.22 &\textbf{60.50} &59.65 &58.01 & & & & & & & \\\cmidrule{2-7}\cmidrule{9-14}
AVG &60.06 &62.98 &64.77 &\textbf{65.36} &64.18 &64.23 & &50.96 &52.44 &55.10 &56.25 &\textbf{57.05} &56.70 & & & & & & & \\
\bottomrule
\end{tabular}
\caption{F1-macro scores across eight languages, comparing our method with the \texttt{Mono} and \texttt{Röttger} baselines. Results are reported for all of target training sizes from 10 to 2,000. Columns represent the number of retrieved instances. \texttt{AVG} denotes the mean over 12 training sizes. The best result for each language and training size is in \textbf{bold}. Retrieved results that outperform the next larger \texttt{Mono} training size are \underline{underlined}.}\label{tab:full_main_table}

\end{table}

Additionally, the underlined values—indicating the smallest amount of retrieved data that outperforms the next larger training size—demonstrate that, in most cases and languages, retrieving cross-lingual data can effectively compensate for having less labeled data. While in two tasks, \textit{Bas19\_es} and \textit{Xdomain\_tr}, training data appears to be of higher quality and retrieval cannot fully offset its absence, in all other tasks, retrieval proves effective especially valuable when labeled data is scarce—less than 50-highlighting the method’s strength in very low-resource hate speech detection settings.

\section{Further Analysis}
\label{sec:full_analysis}

\begin{figure}[htbp]
  \centering
  \begin{subfigure}[b]{0.325\textwidth}
    \includegraphics[width=\textwidth]
    {figures/sankey_tasks.pdf}
  \end{subfigure}
  \begin{subfigure}[b]{0.325\textwidth}
    \includegraphics[width=\textwidth]{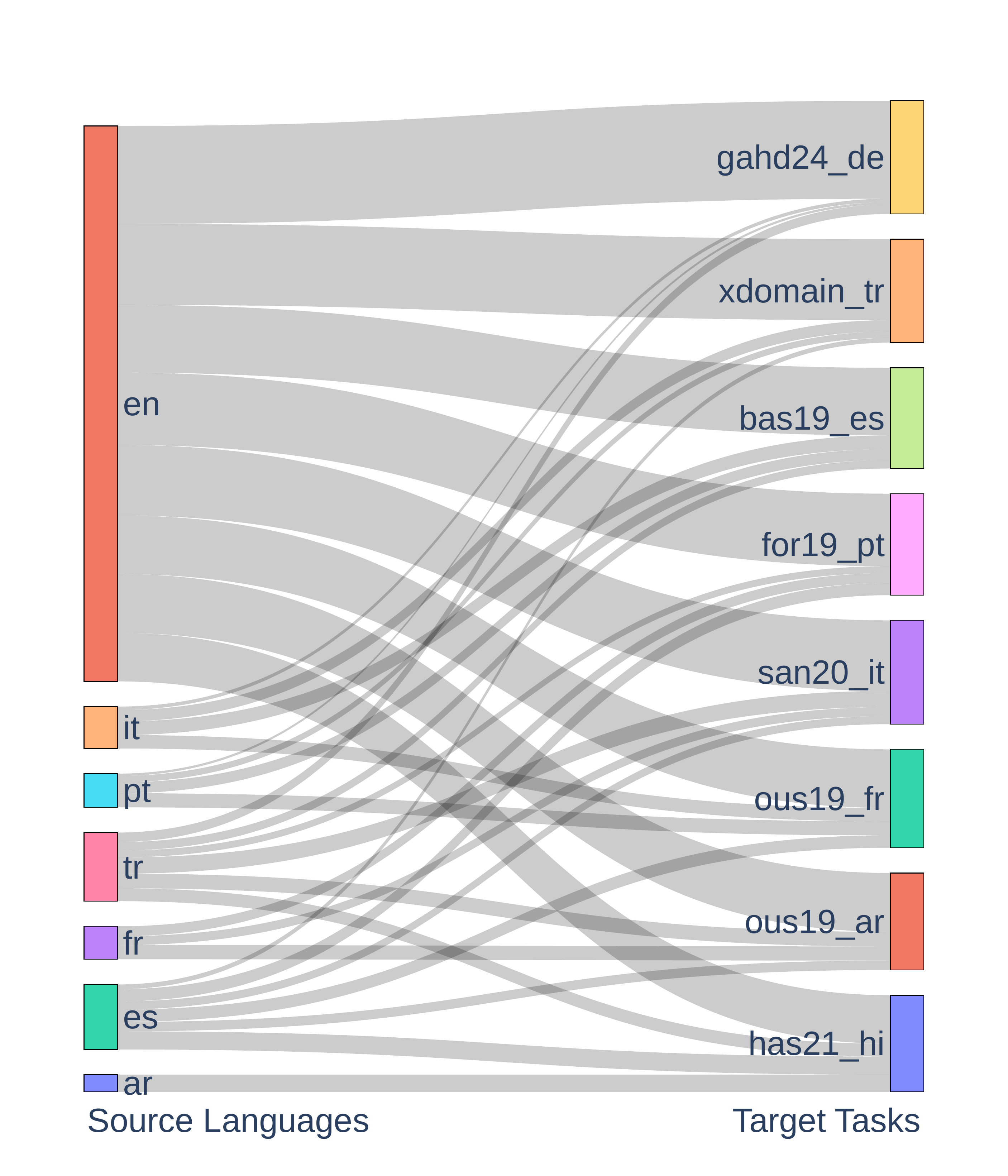}
  \end{subfigure}
  \begin{subfigure}[b]{0.325\textwidth}
    \includegraphics[width=\textwidth]{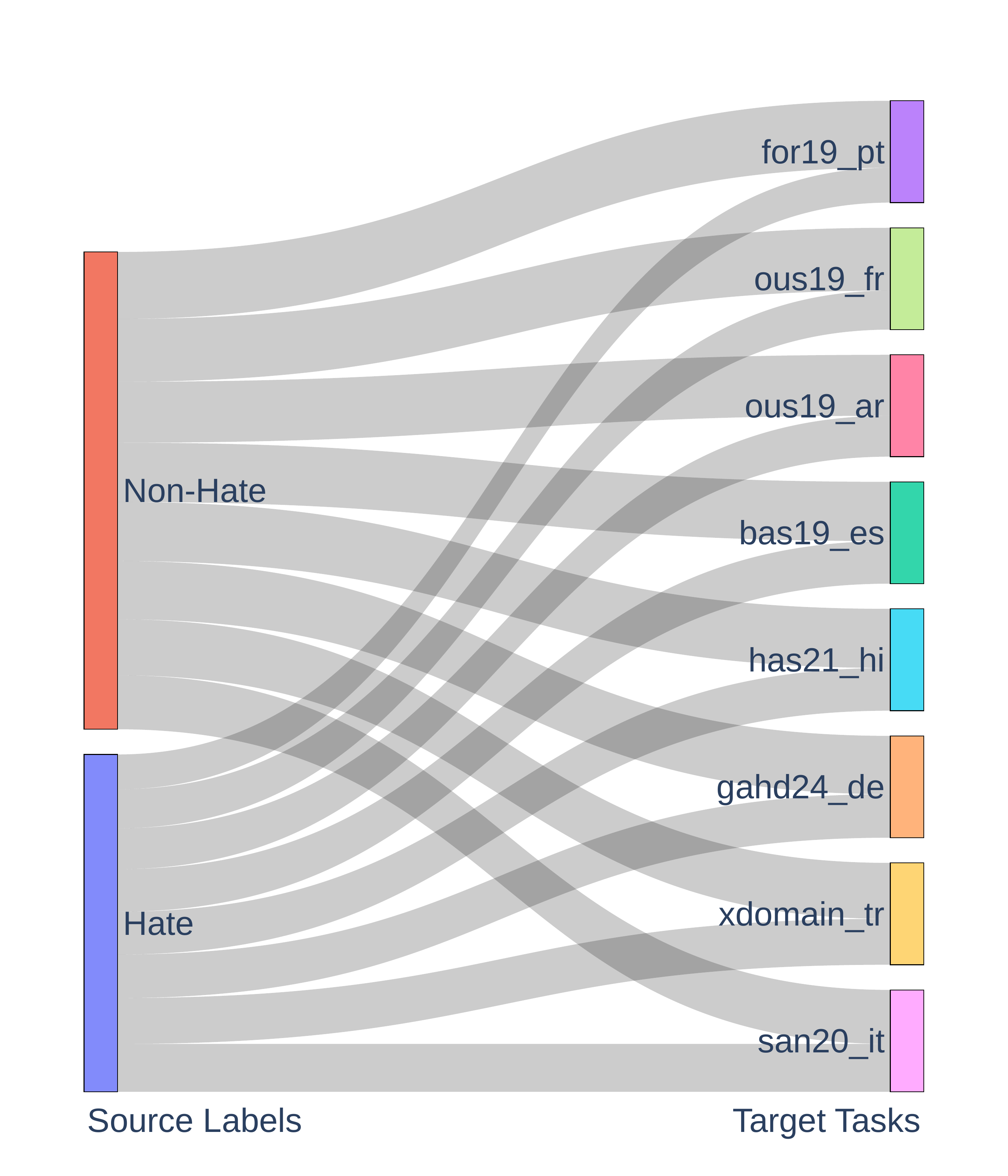}
  \end{subfigure}
  \caption{Sankey diagrams of the distribution of the top four retrieved source tasks (left), languages (middle), and labels (right) for each target task.}
  \label{fig:sankey_2}
\end{figure}

To further analyze what is retrieved and used for fine-tuning under a controlled setting (retrieving 2{,}000 instances for a training size of 2{,}000), see Figure~\ref{fig:sankey_2}. This figure shows the distribution of retrieved source tasks, languages, and labels. As illustrated, English is retrieved the most, followed by Turkish and Spanish in nearly equal amounts, then Italian and Portuguese. Excluding English, this pattern roughly reflects the overall language distribution in the pool (see the "\% of Total Instances" column in Table~\ref{tab:detailed_dataset_information}). The second to fifth most represented languages in the pool are Turkish, German, Italian, and Spanish. However, the low retrieval of German—despite its high presence—and the higher retrieval of Spanish over Italian are unexpected and may be attributed to task generality or cross-lingual similarity.
The distribution of retrieved labels also mirrors their proportions in the pool: approximately 40\% of instances are labeled as hate, and a similar pattern is observed in the retrieved hate instances across target tasks.
 
\section{More about English-only vs. Multilingual Retrieval}
\label{sec:english_vs_multilingual}
Additional results comparing English-only retrieval and multilingual retrieval are presented in Table~\ref{tab:full_english_only_vs_multilingual}. Comparisons in this table should be made vertically: for each language, the values for a specific training size and number of retrieved instances should be compared between the upper (English-only) and lower (multilingual) blocks.
Similar to the datasets discussed in the main text, multilingual retrieval—with even a small number of non-English source tasks—proves beneficial for three languages in this table: \textit{Ous19\_ar}, \textit{San20\_it}, and \textit{Gahd24\_de}, in most cases. However, for \textit{Xdomain\_tr}, the results differ slightly; English-only retrieval performs marginally better, likely due to the generality of English tasks and their semantic similarity to the Turkish dataset.

\begin{table}[!htp]\centering
\scriptsize
\begin{tabular}{p{0.00001cm}p{0.01cm}p{0.5cm}p{0.4cm}p{0.4cm}p{0.4cm}p{0.4cm}p{0.00001cm}p{0.4cm}p{0.4cm}p{0.4cm}p{0.4cm}p{0.00001cm}p{0.4cm}p{0.4cm}p{0.4cm}p{0.4cm}p{0.00001cm}p{0.4cm}p{0.4cm}p{0.4cm}p{0.4cm}}\toprule
& & &\multicolumn{4}{c}{Ous19\_ar} & &\multicolumn{4}{c}{San20\_it} & &\multicolumn{4}{c}{Gahd24\_de} & &\multicolumn{4}{c}{Xdomain\_tr} \\\cmidrule{4-7}\cmidrule{9-12}\cmidrule{14-17}\cmidrule{19-22}
& &SIZE&20 &200 &2000 &20000 & &20 &200 &2000 &20000 & &20 &200 &2000 &20000 & &20 &200 &2000 &20000 \\\midrule
\multirow{6}{*}{\rotatebox{90}{English}} &\multirow{6}{*}{\rotatebox{90}{Retrieval}} &20 &55.75 &60.46 &\textbf{62.22} &\textbf{62.22} & &58.37 &66.30 &67.01 &\textbf{62.23} & &50.30 &55.72 &\textbf{59.79} &54.79 & &62.39 &\textbf{71.47} &\textbf{76.53} &74.01 \\
& &50 &\textbf{64.05} &65.86 &66.34 &55.91 & &\textbf{68.34} &68.53 &62.49 &56.88 & &\textbf{56.96} &60.02 &58.73 &56.62 & &\textbf{76.95} &77.91 &73.85 &57.79 \\
& &200 &67.67 &68.11 &66.86 &\textbf{67.22} & &71.48 &\textbf{73.41} &71.48 &69.73 & &66.88 &66.37 &65.66 &63.72 & &81.84 &82.17 &83.61 &\textbf{83.44} \\
& &500 &\textbf{69.38} &68.88 &\textbf{67.96} &66.81 & &75.19 &74.92 &67.32 &\textbf{73.54} & &\textbf{70.06} &69.81 &63.28 &\textbf{62.12} & &\textbf{85.91} &85.47 &84.50 &75.03 \\
& &2000 &\textbf{70.95} &71.32 &69.26 &\textbf{70.21} & &77.05 &77.75 &68.58 &\textbf{77.69} & &78.93 &78.74 &77.93 &77.58 & &87.81 &87.67 &87.55 &\textbf{87.26} \\\cmidrule{4-7}\cmidrule{9-12}\cmidrule{14-17}\cmidrule{19-22}
& &AVG &\textbf{64.27} &66.46 &66.15 &65.06 & &\textbf{67.74} &\textbf{71.14} &67.91 &66.37 & &62.81 &64.45 &64.79 &62.93 & &\textbf{77.46} &\textbf{80.27} &\textbf{80.90} &76.36 \\\midrule
\multirow{6}{*}{\rotatebox{90}{Multilingual}} &\multirow{6}{*}{\rotatebox{90}{Retrieval}} &20 &\textbf{57.63} &\textbf{63.23} &61.73 &59.47 & &\textbf{63.20} &66.76 &67.06 &60.06 & &50.52 &\textbf{58.15} &59.08 &\textbf{57.48} & &\textbf{66.58} &67.08 &70.14 &\textbf{75.87} \\
& &50 &59.36 &\textbf{66.65} &66.31 &\textbf{64.51} & &67.20 &68.42 &\textbf{71.65} &\textbf{69.44} & &54.53 &60.30 &\textbf{60.47} &\textbf{61.02} & &75.92 &77.50 &\textbf{78.85} &\textbf{70.60} \\
& &200 &67.98 &\textbf{69.18} &67.35 &65.47 & &\textbf{72.46} &72.83 &\textbf{72.41} &\textbf{72.50} & &66.95 &66.15 &\textbf{66.24} &\textbf{65.97} & &81.61 &82.61 &83.04 &82.61 \\
& &500 &68.95 &\textbf{69.47} &69.28 &65.54 & &\textbf{75.39} &\textbf{75.29} &\textbf{74.53} &66.09 & &69.78 &69.68 &\textbf{70.45} &61.80 & &84.93 &85.09 &84.92 &83.88 \\
& &2000 &69.52 &69.77 &\textbf{70.15} &\textbf{68.27} & &\textbf{69.27} &\textbf{78.36} &\textbf{77.57} &76.95 & &\textbf{79.19} &\textbf{78.79} &77.82 &77.90 & &87.39 &\textbf{88.00} &87.39 &\textbf{77.48} \\\cmidrule{4-7}\cmidrule{9-12}\cmidrule{14-17}\cmidrule{19-22}
& &AVG &63.94 &\textbf{66.82} &\textbf{66.41} &\textbf{65.20} & &67.68 &71.00 &\textbf{71.06} &\textbf{68.55} & &\textbf{62.98} &\textbf{64.77} &\textbf{65.36} &\textbf{64.18} & &76.78 &78.88 &79.66 &\textbf{77.58} \\
\bottomrule
\end{tabular}
\caption{F1-macro scores across two strategies: English-only retrieval, and Multilingual retrieval. 
Results are shown for target training sizes of 20, 50, 200, 500, 2{,}000, and \texttt{AVG} (the average over 12 training sizes). Columns represent target languages, and sub-columns are the number of retrieved instances.
}\label{tab:full_english_only_vs_multilingual}
\end{table}

\section{More about Maximum Marginal Relevance}
\label{sec:full_mmr}
To compare the effect of using MMR versus not using it, refer to Table~\ref{tab:mmr_2}. In this table, since each language is presented as a row block, comparisons should be made horizontally within the two sub-tables. Specifically, for each language, values for the same training and retrieval size (e.g., 20 train, 20 retrieved) should be compared between the settings without and with MMR. As shown, for the remaining languages in this table, applying MMR does not lead to significant overall performance differences. This is likely because, even without MMR, the retrieved cross-lingual data is already sufficiently diverse, so applying MMR has minimal additional impact.

\begin{table}[!htp]\centering
\scriptsize
\begin{tabular}{p{0.01cm}p{0.3cm}p{0.35cm}p{0.35cm}p{0.35cm}p{0.35cm}p{0.0000001cm}p{0.35cm}p{0.35cm}p{0.35cm}p{0.35cm}p{0.0000001cm}p{0.01cm}p{0.35cm}p{0.35cm}p{0.35cm}p{0.35cm}p{0.0000001cm}p{0.35cm}p{0.35cm}p{0.35cm}p{0.35cm}}\toprule
& &\multicolumn{4}{c}{Without MMR} & &\multicolumn{4}{c}{With MMR} & & &\multicolumn{4}{c}{Without MMR} & &\multicolumn{4}{c}{With MMR} \\\cmidrule{3-6}\cmidrule{8-11}\cmidrule{14-17}\cmidrule{19-22}
& SIZE &20 &200 &2000 &20000 & &20 &200 &2000 &20000 & & & 20 &200 &2000 &20000 & &20 &200 &2000 &20000 \\\midrule
\multirow{5}{*}{\rotatebox{90}{Bas19\_es}} &20 &54.37 &59.72 &62.52 &63.08 & &54.20 &\textbf{60.15} &62.73 &63.18 & & \multirow{5}{*}{\rotatebox{90}{Ous19\_fr}} &47.21 &52.68 &53.93 &55.05 & &47.25 &52.58 &\textbf{54.31} &55.13 \\
&50 &\textbf{60.93} &64.37 &\textbf{65.59} &\textbf{64.30} & &59.84 &64.85 &64.67 &62.53 & & & 48.29 &52.19 &52.97 &\textbf{55.60} & &48.67 &52.15 &\textbf{54.99} &54.82 \\
&200 &\textbf{72.22} &71.77 &71.23 &\textbf{70.67} & &71.84 &\textbf{72.29} &71.28 &68.80 & & & 51.54 &\textbf{54.06} &55.80 &53.63 & &\textbf{53.70} &53.79 &55.06 &\textbf{54.43} \\
&500 &78.01 &\textbf{77.09} &\textbf{77.79} &67.67 & &77.97 &75.72 &76.64 &\textbf{74.06} & & & \textbf{53.30} &52.84 &\textbf{55.51} &\textbf{55.31} & &51.84 &\textbf{53.13} &54.15 &53.01 \\
&2000 &80.62 &80.50 &80.65 &\textbf{81.02} & &80.31 &\textbf{81.23} &\textbf{81.12} &80.58 & & & \textbf{53.51} &\textbf{53.13} &\textbf{53.30} &\textbf{54.74} & &52.07 &51.44 &52.79 &53.19 \\\cmidrule{3-6}\cmidrule{8-11}\cmidrule{14-17}\cmidrule{19-22}
&AVG &67.27 &69.53 &70.53 &68.69 & &67.26 &\textbf{69.70} &70.01 &68.28 & & & 50.56 &53.12 &54.05 &54.84 & &50.43 &52.99 &54.22 &54.54 \\\midrule
\multirow{5}{*}{\rotatebox{90}{Ous19\_ar}} &20 &57.63 &\textbf{63.23} &\textbf{61.73} &59.47 & &57.75 &59.96 &60.50 &\textbf{62.33} & & \multirow{5}{*}{\rotatebox{90}{For19\_pt}} &49.72 &64.92 &68.57 &\textbf{68.03} & &\textbf{54.31} &\textbf{65.48} &68.95 &67.63 \\
&50 &59.36 &66.65 &\textbf{66.31} &64.51 & &59.66 &66.19 &64.98 &64.46 & & & 59.26 &67.01 &67.06 &\textbf{69.35} & &\textbf{61.81} &67.52 &67.81 &67.38 \\
&200 &67.98 &69.18 &\textbf{67.35} &\textbf{65.47} & &67.61 &69.52 &66.35 &64.55 & & & \textbf{69.69} &\textbf{70.33} &\textbf{70.20} &\textbf{71.07} & &68.29 &69.83 &69.48 &69.74 \\
&500 &68.95 &69.47 &69.28 &65.54 & &68.70 &\textbf{70.39} &69.59 &\textbf{67.17} & & & 69.72 &70.84 &\textbf{70.04} &71.05 & &69.46 &70.21 &69.91 &71.83 \\
&2000 &69.52 &69.77 &\textbf{70.15} &68.27 & &\textbf{70.20} &\textbf{70.12} &69.93 &68.19 & & & 72.39 &\textbf{72.66} &71.72 &72.22 & &72.24 &70.85 &71.93 &72.48 \\\cmidrule{3-6}\cmidrule{8-11}\cmidrule{14-17}\cmidrule{19-22}
&AVG &63.94 &66.82 &\textbf{66.41} &65.20 & &\textbf{64.16} &66.96 &65.81 &64.94 & & & 62.85 &68.18 &69.55 &69.69 & &\textbf{63.60} &68.28 &69.28 &69.75 \\\midrule
\multirow{5}{*}{\rotatebox{90}{Has21\_hi}} &20 &47.34 &51.03 &53.68 &55.37 & &47.42 &\textbf{52.96} &53.17 &56.03 & & & & & & & & & & & \\
&50 &\textbf{48.39} &53.36 &52.26 &55.78 & &47.34 &\textbf{54.19} &\textbf{53.55} &\textbf{56.29} & & & & & & & & & & & \\
&200 &55.83 &54.65 &56.80 &56.02 & &55.57 &\textbf{56.11} &56.67 &\textbf{57.67} & & & & & & & & & & \\
&500 &56.94 &57.66 &57.88 &\textbf{59.55} & &56.78 &\textbf{58.43} &\textbf{59.28} &57.55 & & & & & & & & & & & \\
&2000 &\textbf{58.19} &60.22 &60.50 &59.65 & &57.37 &60.02 &60.05 &\textbf{61.10} & & & & & & & & & & & \\\cmidrule{3-6}\cmidrule{8-11}
&AVG &52.44 &55.10 &56.25 &57.05 & &52.38 &55.22 &56.08 &57.37 & & & & & & & & & & & \\
\bottomrule
\end{tabular}
\caption{F1-macro scores without/with MMR for five languages (rows), across five selected training sizes and an average (\texttt{AVG}) computed over 12 training sizes.
}\label{tab:mmr_2}
\end{table}

\end{document}